\documentclass[12pt,letterpaper,onecolumn]{autart}
\usepackage{graphicx}          
\usepackage[dvips]{epsfig}    
\usepackage{cite}
\usepackage{amssymb}
\usepackage{amsmath}
\usepackage{bm}

\usepackage{graphicx}
\usepackage{epstopdf}
\usepackage{subfigure}
\usepackage{color}
\usepackage{algorithm}
\usepackage{algorithmicx,algpseudocode}

\usepackage{graphics}
\usepackage{booktabs}
\usepackage{array}
\usepackage{multirow}
\usepackage{url}

\DeclareMathOperator*{\tr}{{\rm \tr}}

\newtheorem{definition}{Definition}

\begin{document}

\begin{frontmatter}
\title{Branch and Bound in Mixed Integer Linear Programming Problems: A Survey of Techniques and Trends}

\author[HKUST]{Lingying Huang}\footnotemark[1]\ead{lhuangaq@connect.ust.hk},
\author[HKUST]{Xiaomeng Chen}\footnotemark[1]\ead{xiaomeng.chen@connect.ust.hk},
\author[HKUST]{Wei Huo}\footnotemark[1]\ead{whuoaa@connect.ust.hk},
\author[HW]{Jiazheng Wang}\ead{wang.jiazheng@huawei.com},
\author[HW]{Fan Zhang}\ead{zhang.fan2@huawei.com},
\author[HW]{Bo Bai}\ead{baibo8@huawei.com},
\author[HKUST]{Ling Shi}\ead{eesling@ust.hk}
\address[HKUST]{Department of Electronic Engineering, HKUST, Clear Water Bay, Kowloon, Hong Kong}
\address[HW]{Theory Lab, Huawei Hong Kong Research Centre, Hong Kong SAR, China}
\footnotetext[1]{These authors contributed equally to this work.} 

\begin{abstract}
In this paper, we surveyed the existing literature studying different approaches and algorithms for the four critical components in the general branch and bound (B\&B) algorithm, namely, branching variable selection, node selection, node pruning, and cutting-plane selection. However, the complexity of the B\&B algorithm always grows exponentially with respect to the increase of the decision variable dimensions. In order to improve the speed of B\&B algorithms, learning techniques have been introduced in this algorithm recently. We further surveyed how machine learning can be used to improve the four critical components in B\&B algorithms. In general, a supervised learning method helps to generate a policy that mimics an expert but significantly improves the speed. An unsupervised learning method helps choose different methods based on the features. In addition, models trained with reinforcement learning can beat the expert policy, given enough training and a supervised initialization. Detailed comparisons between different algorithms have been summarized in our survey.  Finally, we discussed some future research directions to accelerate and improve the algorithms further in the literature.		
\end{abstract}


\end{frontmatter}

\section{Introduction}
Branch and bound (B\&B) algorithm is a widely-used method to produce exact solutions to non-convex  and combinatorial problems which cannot be solved in polynomial time. It was initially proposed by Land and Doig  \cite{land2010automatic} to solve discrete programming problems.  This method implicitly enumerates all possible solutions by iteratively dividing the original problem into a series of sub-problems, organized in a tree structure, and discarding the sub-problems where a global optimum cannot be found. The approaches adopted to generate the sub-problems of the unexplored nodes in the tree represent the ``branching'' step, while the ``bounding'' phase consists of rules used to prune off regions of sub-optimal search space. Once the entire tree has been explored, the exact solution can be achieved. 

Many decisions affect the performance of B\&B by guiding the search to promising space and enhancing  the chance of quickly finding an exact solution. These decisions are the \textbf{variable selection} (i.e., which of the fractional variables  to branch on), the \textbf{node selection} (i.e., which of the current nodes to explore next), the \textbf{pruning rules} (i.e., rules that prevent exploration of the sub-optimal space) and the \textbf{cutting rules} (i.e., rules that add constraints to find cutting planes). A well-designed decision strategy can reduce the search space and thus speed up the search progress of the B\&B algorithm significantly.

Substantial  research has been done to study the efficient and manually-designed strategies to deal with the four decisions \cite{clausen1999branch},\cite{morrison2016branch}. Unfortunately, the traditional strategies are only designed for each problem type, and are not adopted to a family of problems. In order to speed up B\&B  that are adapted to a family of problems, machine learning components are integrated.  The aim of this paper is to provide a survey of  learning-based B\&B, particularly regarding the above four decisions. 

\subsection{Motivation}
For a B\&B algorithm, machine learning improves its performance on a family of problems in two ways. On the one hand, expert knowledge is assumed about the decision strategy, while some heavy computation is preferred to be substituted by an approximation method. Under this circumstance, the learning method can generate such approximation without the need  to devise new algorithms. On the other hand, the expert knowledge may be insufficient, which cannot satisfy some decisions. Hence, in this case,  the learning approach aims to explore the space of all decisions and learn the best performing policy from this experience, improving the algorithm performance.  For a  machine learning method, B\&B is able to decompose the original problem into smaller learning tasks. The tree structure of  B\&B acts as a relevant prior for the learning model.

\subsection{Setting}
A common setting where B\&B is adopted is the Mixed-Integer Linear Programming (MILP) problem, which is defined as follows.

\begin{definition}
	\textbf{(MILP)}. Given a matrix $A\in \mathbb{R}^{m\times n}$, vectors $b\in \mathbb{R}^{m}$ and $c \in \mathbb{R}^{n}$, and a subset $I \subseteq \{1,\dots, n\},$ the mixed-integer linear program $MILP=(A,b,c,I)$ is:
	\begin{equation*}
	z^\star= \min \{c^T x | Ax\leq b, x\in \mathbb{R}^n, x_j\in \mathbb{Z},\forall j\in I \}
	\end{equation*}
\end{definition}
The vectors in the $X_{MILP}=\{x\in \mathbb{R}^n |Ax\leq b, x\in \mathbb{R}^n, x_j\in \mathbb{Z},\forall j\in I\}$ are called \textit{feasible solution} of MILP. A feasible solution $x^\star \in X_{MILP}$ of MILP is  \textit{optimal} if its objective value satisfies $c^T x^\star=z^\star$.

Owing to the integrality requirement, MILP is an NP-hard problem. Most modern MILP solvers such as CPLEX\cite{CPLEX}, LINDO\cite{LINDO} and SIP\cite{martin1999integer}, iteratively split the problem into smaller subproblems, building a search tree (see Section \ref{bb}). To solve MILP, B\&B implements a divide-and-conquer algorithm, where a linear programming (LP) relaxation of the problem is computed by removing the integrality conditions. 
\begin{definition}
	\textbf{(LP relaxation of a MILP)}. The LP  relaxation of a MILP is:
	\begin{equation*}
	\check z= \min \{c^T x | Ax\leq b, x\in \mathbb{R}^n \}
	\end{equation*}
\end{definition}

A lower bound for the whole subtree is provided by solving the LP relaxation, and if the objective value $\check z$ of the LP relaxation is larger than or equal to the value $\hat z=c^T\hat x$ of the current best solution $\hat x$, the corresponding node  can be discarded. 

The most essential components  of  an MILP solver implementation are a branching rule, a node selection strategy, a fast and stable LP solver and cutting plane separators, which will be introduced in the next section.

\subsection{Outline}
We have introduced the context and motivations for building the B\&B  algorithms together with machine learning. The remainder of this survey is organized as follows. Section \ref{s2} overviews the implementation of the general B\&B algorithm and summarizes the existing literature studying different approaches and algorithms for the four key components, namely, branching variable selection, node selection, node pruning, and cutting-plane selection. Section \ref{s3} provides prerequisites in machine learning to make the reader familiar with some essential concepts required to understand the learning-based B\&B algorithms. Section \ref{s4} provides a survey of the learning techniques to deal with the four critical components in B\&B algorithms for MILP. Further elaboration on the contributions and limitations of different studies are provided. Section \ref{s5} presents some future work directions, and Section \ref{s6} summarizes the contributions of this survey.

\section{Branch and Bound Algorithms}\label{s2}
\begin{figure}
	\centering
	\includegraphics[width=\textwidth]{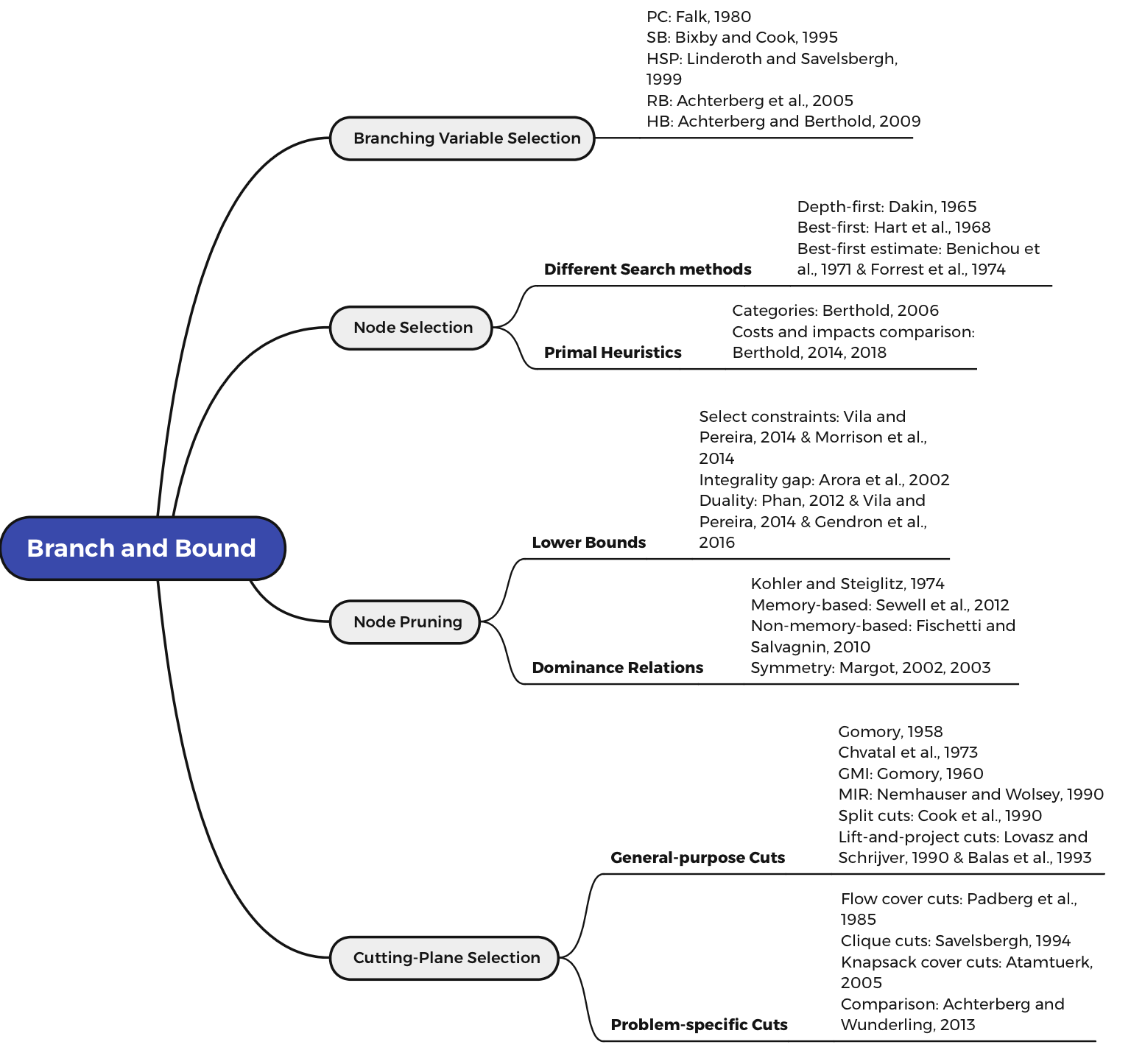}
	\caption{Four core components in B\&B algorithms and their related work.}
	\label{BB}
\end{figure}

In this section, we give an overview of the general B\&B algorithm, along with a detailed description of the four core  components in B\&B. The main survey results are summarized in Figure \ref{BB}.
\subsection{Algorithm Overview}\label{bb}
Define an optimization problem as $\mathcal{P}=(\mathcal{D}, f)$, where $\mathcal{D}$ (\textbf{search space}) is denoted as a set of valid solutions to the problem and $f: \mathcal{D} \rightarrow \mathbb{R}$ is denoted as the \textbf{objective function}. The problem  $\mathcal{P}$ aims to find an optimal solution $x^\star \in \arg\min_{x\in\mathcal{D}} f(x)$. A \textbf{search tree} $T$  of \textbf{subproblems} is built
by   Branch and-Bound in order to solve problem  $\mathcal{P}$. Moreover, a feasible solution $\hat x \in \mathcal{D}$ is globally stored. At each iteration, B\&B selects a new subset of the search space $\mathcal{S} \subset \mathcal{D}$ for exploration from a queue $\mathcal{L}$ of unexplored subsets. Then, if a solution $\hat x' \in \mathcal{S}$ (\textbf{candidate incumbent}) has a better objective value than $\hat x$, i.e., $f(\hat x')<f(\hat x)$, the incumbent solution is updated. On the other side, the subset is \textbf{pruned} or \textbf{fathomed} if there is no solution in $\mathcal{S}$ with better objective solution than $\hat x$, i.e., $f(\hat x)\geq f(\hat x),\forall x \in \mathcal{S}$. Otherwise, the subset  $\mathcal{S}$ is branched into child subproblems $\mathcal{S}_1, \mathcal{S}_2, \ldots, \mathcal{S}_r$, which are then pushed onto $\mathcal{L}$. Once there is no unexplored subsets in queue $\mathcal{L}$, the best incumbent solution is returned and the algorithm terminates. Pseudocode for the generic B\&B is given in  Algorithm \ref{agl1}. 
\begin{algorithm}[htp]
	\caption{Branch and Bound $(\mathcal{D}, f)$} \label{agl1}
	\begin{algorithmic}[1]
		\State Set $\mathcal{L}=\mathcal{D}$ and initialize $\hat x$
		\While{$\mathcal{L}\neq\emptyset$} 
		\State Select a subproblem $\mathcal{S}$ from $\mathcal{L}$ to explore
		\If{a solution $\hat x' \in\{x\in \mathcal{S}| f(x)<f(\hat x)\}$ can be found} 
		\State Set $\hat x =\hat x'$ 
		\EndIf
		\If{$\mathcal{S}$ cannot be pruned} 
		\State Partition $\mathcal{S}$ into $\mathcal{S}_1, \mathcal{S}_2, \ldots, \mathcal{S}_r$
		\State Insert $\mathcal{S}_1, \mathcal{S}_2, \ldots, \mathcal{S}_r$ into $\mathcal{L}$
		\EndIf
		\State Remove $\mathcal{S}$ from $\mathcal{L}$
		\EndWhile
		\State \Return $\hat x$
	\end{algorithmic}
\end{algorithm}

In terms of the above pseudocode, the variable selection strategy (branching rules) affects how the subproblem is partitioned in Line 7 of Algorithm \ref{agl1}; the node selection strategy affects the order of which nodes is selected to explore (Line 3 of Algorithm \ref{agl1}), and the pruning rule in Line 6 of Algorithm \ref{agl1} determines if $\mathcal{S}$ is fathomed.

\begin{figure}[ht]
	\centering
	\includegraphics[width=\textwidth]{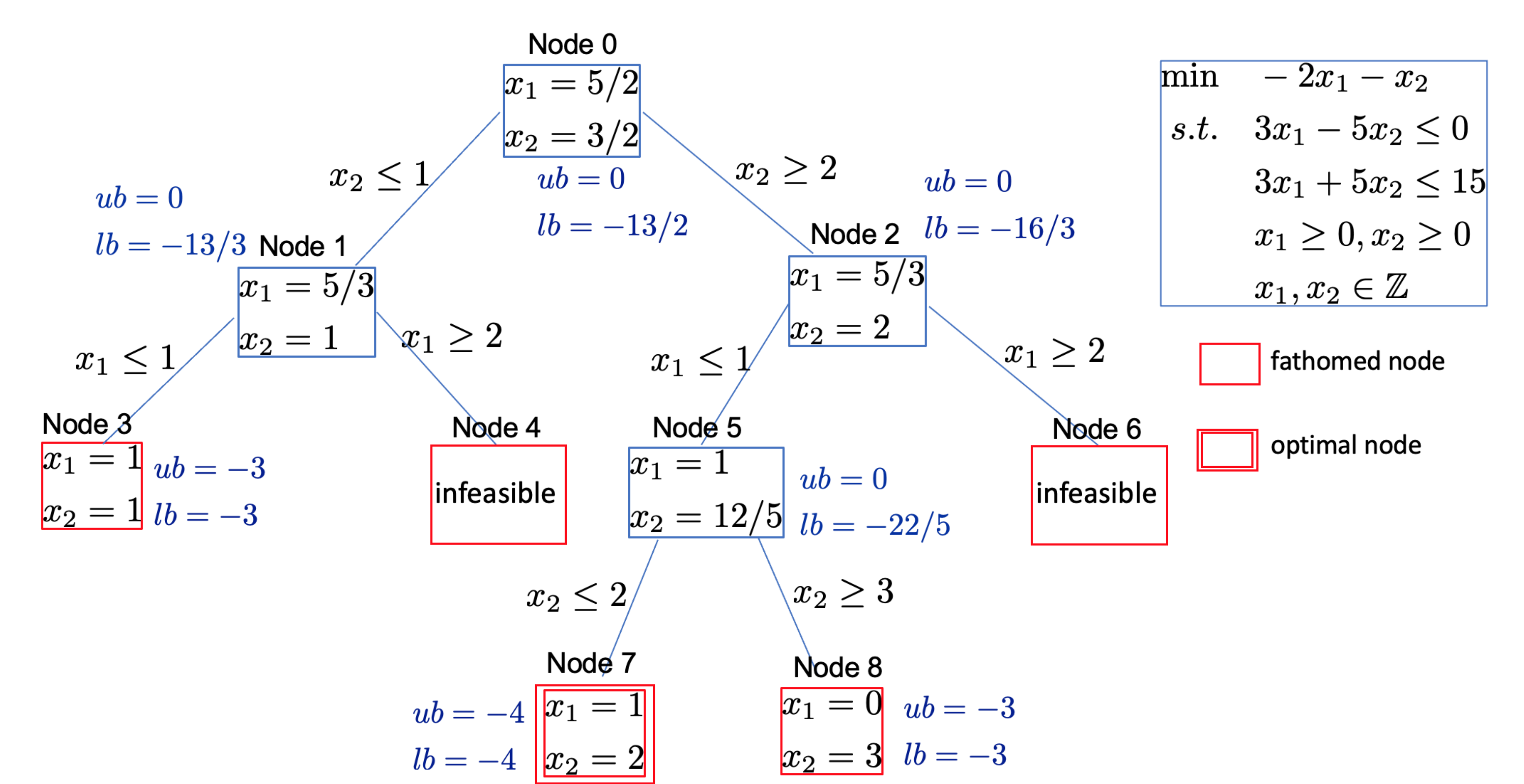}
	\caption{Adopting B\&B to solve a minimization integer linear programming.}
	\label{f1}
\end{figure}

Figure \ref{f1} illustrates a concrete example of B\&B algorithm for a minimization mixed linear integer programming. The optimization problem is shown in the upper right corner of  the Figure \ref{f1}. The original upper bound is $0$, which is computed at  $x_1=0,x_2=0$. At each node, a local lower bound is computed by the LP relaxation problem by the LP solver. A local upper bound  is  updated when an integer solution is found. At each iteration, we can compare the solution of the current LP relaxation  with the minimum upper bound so far, which is called the incumbent solution value. If the latter is smaller than the former for a certain subproblem, no better mixed-integer solution can be found and the node can be fathomed. In Figure \ref{f1}, the fathomed nodes are shown in red rectangles.

In order to enhance the above enumeration framework, mixed-integer linear programming solvers tend to adopt the cutting planes (linear inequalities) to the original formulation, especially at the root node. By adding cuts to the original linear programming, some parts of the feasible region is eliminated which strengthens the LP relaxation.  

\subsection{Branching Variable Selection} \label{BVS}
As a critical task in Brand-and-Bound, branching variable selection decides how to partition a current node into two-child nodes recursively. Specifically, it decides which fractional variables (also known as candidates) to branch on. Branching on an inefficient variable that does not simplify subproblems doubles the size of the B\&B tree, thus making no sense. The ultimate objective of an effective branching strategy is to minimize the number of explored nodes before the algorithm terminates. To indicate the quality of a candidate variable, \emph{score} of this variable is used to measure its effectiveness, and the candidate with the highest score is picked to branch on. Pseudocode for the generic variable selection is presented in Algorithm~\ref{alg: VB}.
\begin{algorithm}[t]
	\caption{Branching Variable Selection} \label{alg: VB}
	\hspace*{0.02in} {\bf Input:}
	Subproblem of the current node $\mathcal{S}$ with its optimal LP solution 
	$\hat{x} \notin X_{MILP}$\\
	\hspace*{0.02in} {\bf Output:} 
	A subscript $i \in I$ of an integer variable with fractional value 
	$\hat{x}_{i} \notin \mathbb{Z}$
	\begin{algorithmic}[1]
		\State Define branching candidates set 
		$C = \{ i\in I \mid \hat{x}_{i} \notin \mathbb{Z} \}$
		\State For each candidate $i \in C$, calculate its score value $s_{i} \in \mathbb{R}$
		\State \Return $i = \mathop{\arg\min}_{i \in C} \ \ s_{i}$
	\end{algorithmic}
\end{algorithm}

The difference among various branching policies is how the score is computed, and we will introduce different variable selection methods in the following. Making high-quality branching decisions is usually nontrivial and time-consuming. Although a good branching method should produce trees as small as possible, the primary goal of solving large-scale optimizations is to spend as little time as possible. Hence, great branching strategies should compromise between the quality of decisions taken and the time spent taking each decision. This trade-off is the focus of the entire branching study.

An early intuitive strategy is known as \emph{most infeasible branching} (MIB), choosing the most fractional variable to branch on, i.e., branching on a variable whose fractional part is closest to $0.5$. Although this method can be evaluated easily, numerical results in Achterberg et al.~\cite{achterberg2007constraint, ACHTERBERG200542} have indicated that the performance of this rule is worse than that of selecting the variable randomly. Later, \emph{pseudocost branching} (PC) was proposed (see Falk~\cite{Falk1980Experiments}). This method maintaining a history of variables' branchings and corresponding dual bound increases averages previous improvements to obtain the expected gain for each candidate. This expected gain is the score for each candidate variable; thus, the variable that can yield a significant change in the objective value is chosen to branch on. While PC works well in saving computation time, it is inefficient at the beginning of the B\&B tree since there is no reliable history at the root. Hence, the pseudocosts for each variable should be initialized, which requires significant manual tuning. An extreme strategy is \emph{strong branching} (SB), which was first proposed in the context of the traveling salesman problem, shown in Bixby and Cook~\cite{1995FINDING}. This policy chooses the candidate that yields the best improvement in the dual bound. Its full version, \emph{full strong branching} (FSB), solves the resulting LPs optimally for each candidate. This method makes an excellent decision at each step, producing the smallest B\&B tree, but the computation time is extremely high, causing it intractable in practice. Hence, SB tries to accelerate the FSB by only focusing on a smaller candidate set of variables instead of all candidates, and operating fewer simplex pivots rather than solving subproblems optimally. To circumvent the drawbacks of PC and SB, \emph{hybrid strong and pseudocost} (HSB) applies accurate SB in the upper level of the tree until a certain depth and then employs PC based on the initialization given by SB. In this way, the difficulty of PC initialization and expensive computation of SB are both overcome. This idea was first proposed by Falk~\cite{Falk1980Experiments} and further developed by Linderoth and Savelsbergh~\cite{developedSBPC}. More recently, \emph{reliability branching} (RB) further improves the combination of pseudocost and strong branchings, presented in~\cite{ACHTERBERG200542}. It switches between PC and SB according to the ``reliability" of a candidate based on a pre-defined reliability threshold. For a variable, if the number of times it is selected is less than the threshold, it will be considered unreliable, and its score will be calculated via SB. Otherwise, if this candidate has been selected several times, it will be regarded as very reliable. Therefore, we can use the less accurate but very time-efficient pseudocost method to get its score. The last rule worth mentioned here is \emph{hybrid branching} (HB) from Achterberg and Berthold~\cite{HB}, which integrates five kinds of scores from different selection criteria. These scores are normalized and merged into a single value through a weighted sum. In addition to what we have introduced above, there are some other branching strategies, such as \emph{backdoor} branching developed by Fischetti and Monaci~\cite{backdoor} selecting a subset of top-ranked candidates, information-theoretic \emph{entropy branching} in Gilpin and Sandholm~\cite{GILPIN2011147} regarding fractional variables as probabilities and trying to remove as much uncertainty as possible from subproblems by branching, and \emph{nonchimerical branching} shown in Fischetti and Monaci~\cite{FISCHETTI2012159}.  

\subsection{Node Selection}
After a sub-problem has been produced by constraining some integer variables in MILP, the solving process can continue with any sub-problem that is a leaf of the current search tree. We refer to the sub-problems as nodes, and node selection designs which node to process in the next step. The existing literature always selects the next node based on the following two usually opposing goals:
\begin{enumerate}
	\item Finding good feasible MILP solutions to improve the primal (upper) bound, which helps to prune the search tree by bounding;
	\item Improving the global dual (lower) bound.
\end{enumerate}
Therefore,  in the following, we first take a brief overview of different search methods in the literature and then introduce different methods of primal heuristics to find feasible solutions.
\subsubsection{Different Search Methods}
Dakin \cite{dakin1965tree} proposed depth-first search for MILP. This node selection rule always chooses a node from the leaf queue with the maximal depth in its search tree. Depth-first search is the preferred strategy for pure feasibility problems. The sub-problem management is reduced to the minimum due to the similarity between two subsequent sub-problems. Compared with other search methods, it has small memory consumption. However, depth-first search completely disregards the second goal to improve the global dual bound. To improve the global dual bound as fast as possible, the best-first search, which selects a leaf with the currently smallest dual objective value, was introduced by Hart \cite{hart1968formal}. It was shown by Achterberg~\cite{achterberg2007constraint} that there exists a node selection strategy of the best-first search type, which solves the instance in a minimal number of nodes. None of the above methods, however, tries  to improve the quality of integer feasible solution. In order to find a good feasible solution as soon as possible, the best search estimate was proposed. There are mainly two kinds of estimate schemes, i.e., the best projection criterion by B\'{e}nichou et al.~\cite{benichou1971experiments}, and the best estimate rule by Forrest et al.~\cite{forrest1974practical}, which differ in how they determine an estimate of the best solution obtainable from a node. The best projection criterion calculates the objective value increase per unit decrease in infeasibility, while the best estimate rule employs the pseudocost values~\cite{benichou1971experiments} to estimate the increase in the objective value. Linderoth and Savelsbergh~\cite{linderoth1999computational} further indicated that the best estimate rule outperforms the best projection rule in computing time and shows the backtracking methods to improve the node selection method by using estimates to avoid superfluous nodes.

\subsubsection{Primal Heuristics} \label{s2.3.2}
Heuristics are procedures that try to find good solutions in a short time by orientating themselves on some information which is helpful to lead to the desired result. Since it is a costly method and has a worst-case runtime to solve the MILP problems, primal heuristics are crucial to finding quality feasible solutions quickly \cite{berthold2006primal}. In addition, primal heuristics which help find a feasible solution have the following advantages~\cite{achterberg2007constraint}
\begin{enumerate}
	\item  It proves the model is feasible, which is an indication that there is no error in the model.
	\item The quality of the heuristic solution can be adjusted to the user's requirement; thus, the process can be terminated at an early stage.
	\item Feasible solutions help to prune the search tree by bounding which reduces the work of the B\&B algorithm.
	\item The better the current incumbent is, the more reduced cost fixing and other dual reductions can be applied to tighten the problem formulation.
\end{enumerate}

Berthold \cite{berthold2006primal} summarized five classic primal heuristic methods, including OCTANE by Balas et al. \cite{balas2001octane}, feasibility pump by Fischetti et al. \cite{fischetti2005feasibility}, local branching by Fischetti and Lodi \cite{fischetti2003local}, relaxation induced neighborhood search~(RINS) by Danna et al. \cite{danna2005exploring} and Mutation by Rothberg \cite{rothberg2007evolutionary}. They then developed two new methods based on the above classic methods, namely, relaxation enforced neighborhood search~(RENS) and Crossover. The above primal heuristics are grouped into two categories based on whether the algorithm needs a previous feasible point:
\begin{enumerate}
	\item  Start heuristics: find a feasible solution early, such as OCTANE\cite{balas2001octane}, feasibility pump\cite{fischetti2005feasibility} and RENS \cite{berthold2006primal}.
	\item Improve heuristics: start with a primal feasible point and improve either the feasibility condition or the optimality quality iteratively, such as local branching \cite{fischetti2003local}, RINS\cite{danna2005exploring} and Mutation \cite{rothberg2007evolutionary}.
\end{enumerate}

Berthold also summarized the advantages and disadvantages of different primal heuristics in \cite{berthold2006primal}. More overviews comparing different primal heuristics about their computational costs and their impact on problem-solving can be seen in \cite{berthold2014primal,berthold2018computational}.






\subsubsection{Evaluation Criteria}\label{s2.3.3}
To evaluate the quality of a node search criteria, we divide the evaluation criteria into two groups, one related to the feasibility quality, while the other is related to the quality of the feasible solution found so far.

The feasibility quality is evaluated by the pseudocost values of the variables in the best estimate rule of Forrest et al. \cite{forrest1974practical}, i.e.,
\begin{equation}
e_{F}=\sum\limits_{j\in I} \min\{P_{j}^{-}f_{j}^{-},P_{j}^{+}f_{j}^{+}\},
\end{equation}
with $ f_{j}^{-}=x_{j}-\lfloor x_{j}\rfloor $ and $ f_{j}^{+}=\lceil x_{j}\rceil-x_{j}$, where $ x_{j} $ is the current LP solution, $\lfloor\cdot\rfloor$ and $\lceil\cdot\rceil$ are the floor and ceil function, respectively, and the different pseudocoast values initialization can be seen from \cite{linderoth1999computational}.

In order to evaluate the quality of a feasible solution concerning the objective function, three relative gaps are introduced.
\begin{definition} \label{d3}
	Let $ x(t) $ be a feasible solution of a given MILP, $ x^{*} $ be an optimal or best known solution of this MILP at time $ t $, and $ c^{\top}y(t) $ be the current dual bound during a MILP solving process.
	
	The primal gap is defined as
	\begin{equation}
	\gamma_{p}(t)=\left\lbrace \begin{array}{ll}
	1&\text{if } c^{\top}x^{*}\cdot c^{\top}x(t)<0,\\ 
	\dfrac{c^{\top}x(t)-c^{\top}x^{*}}{\max\{|c^{\top}x(t)|,|c^{\top}x^{*}|,\varepsilon\}}& \text{otherwise.}
	\end{array} \right. 
	\end{equation}
	where use $ \varepsilon=10^{-12} $ to avoid the division by 0.
	
	The dual gap is defined as
	\begin{equation}
	\gamma_{d}(t)=\left\lbrace \begin{array}{ll}
	1&\text{if } c^{\top}x^{*}\cdot c^{\top}y(t)<0,\\ 
	\dfrac{c^{\top}x^{*}-c^{\top}y(t)}{\max\{|c^{\top}x^{*}|,|c^{\top}y(t)|,\varepsilon\}}& \text{otherwise.}
	\end{array} \right. 
	\end{equation}
	
	The primal-dual gap is 
	\begin{equation}
	\gamma_{pd}(t)=\left\lbrace \begin{array}{ll}
	1&\text{if } c^{\top}x(t)\cdot c^{\top}y(t)<0,\\ 
	\dfrac{c^{\top}x(t)-c^{\top}y(t)}{\max\{|c^{\top}x(t)|,|c^{\top}y(t)|,\varepsilon\}}& \text{otherwise.}
	\end{array} \right. 
	\end{equation}
	
\end{definition}

To take into account the whole solution process, Berthold \cite{berthold2013measuring} introduced a new performance measure, in particular for benchmarking primal heuristics. The progress of the primal bound's convergence towards the optimal solution over the entire solving time is revealed in this measurement. This method makes use of the primal gap and computes the integral of the function over time. Therefore, we call this measurement the primal integral.
\begin{definition}[Primal Integral]
	Let $ t_{max}\in\mathbb{R}_{\geq0} $ be a limit on the solution time of the B\&B MILP solver. The primal gap function $ p:[0,t_{max}] \longmapsto[0,1]$ is defined as
	\begin{equation}
	p(t)=\left\lbrace \begin{array}{ll}
	1&\text{if no incumbent is found until time }t,\\ 
	\gamma_{p}(t)& \text{otherwise.}
	\end{array} \right. 
	\end{equation}
	
	The primal integral $ P(T) $, $ T\in[0,t_{max}] $ of a B\&B run  is defined as
	\begin{equation}
	P(T)=\int_{t=0}^{T}p(t)dt.
	\end{equation}
\end{definition}

\subsection{Node Pruning} \label{s2.4}
Pruning suboptimal branches is an important part of B\&B algorithms since it keeps the B\&B tree and the computing steps small, which reduces the solving time and the required memory. In a standard B\&B algorithm, the pruning policy prunes a node only if one of the following conditions is met:
\begin{enumerate}
	\item Prune by bound: compute a lower bound on the objective function value at each node. If the lower bound of the node is larger than the optimal objective value obtained, the node will be pruned, i.e., $\check{z}>\hat{z}$.
	\item Prune by infeasibility: if the relaxed problem of this node is infeasible, which means the lower bound of this node is $ \infty $. This can be viewed as a special case of prune by bound.
	\item Prune by integrality: if the obtained solution for the relaxed problem satisfies the integer constraints, it is unnecessary to search the children of this node.
\end{enumerate}
We call the nodes satisfying one of the above conditions as fathomed nodes.

\subsubsection{Lower Bounds}
The most common way to prune is pruning by bound. If the lower bound of the objective function value $\hat{z}$ is smaller, more subproblems can be pruned since the chance for $\check{z}>\hat{z}$ is larger. In general, many different lower bounds can be computed as necessary; some lower bound computations may be easy to perform, whereas others may be more computationally intensive. Vil\`{a} and Pereira \cite{vila2014branch}, and Morrison et al. \cite{morrison2014application} attempted to prune using the manageable lower bounds first and then move on to more complex, but tighter, lower bounds. To improve lower bounds, Arora et al. \cite{arora2002proving} derived a new formulation with a tighter integrality gap.

Another method for deriving lower bounds is through duality. Integer programming duality methods is adopted not limited in Phan \cite{phan2012lagrangian}, Vil\`{a} and Pereira \cite{vila2014branch}, and Gendron et al. \cite{gendron2016lagrangian}.

\subsubsection{Dominance Relations}
Kohler and Steiglitz \cite{kohler1974characterization} first studied the dominance relations. If a node $ S_{1} $ dominates the other node $ S_{2} $, this means that for any solution that is a descendant of  $ S_{2} $, there exists a complete solution descending from $ S_{1} $ which is no worse than that. Therefore, dominance relations allow nodes to be pruned if they are dominated by some other nodes. There are two primary types of dominance relations, the memory-based dominance rules such as Sewell et al. \cite{sewell2012bb} and the non-memory-based dominance rules such as Fischetti and Salvagnin \cite{fischetti2010pruning}. The memory-based dominance rules require the entire search tree to be stored for the duration of the algorithm. In contrast, non-memory-based dominance relations do not require the dominating state to have been previously generated in the search process. For example, by solving an auxiliary problem, the method in Fischetti and Salvagnin \cite{fischetti2010pruning} is able to identify whether a node in the tree is dominated by some other node regardless of whether explored before. Note that Demeulemeester et al. \cite{demeulemeester2000discrete} showed that at least one node cannot be pruned if there is a dominance cycle. Ibaraki \cite{ibaraki1977power} showed that dominance relations do not always improve the pruning quality; however, dominance relations would prune a lot of redundant nodes for problems with a high degree of symmetry. Margot \cite{margot2002pruning,margot2003exploiting} introduced isomorphic pruning to recognize symmetry.

\subsection{Cutting-Plane Selection}
\begin{figure}[htbp]  
	\centering
	\begin{tabular}{cc}
		\subfigure[Feasible region of original LP relaxation]{
			\includegraphics[width=0.3\linewidth]{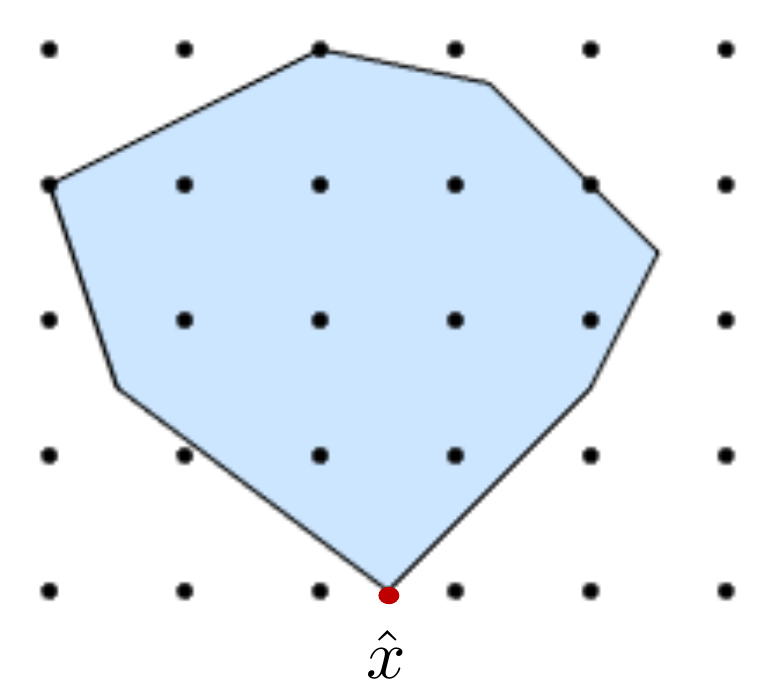}}  &
		\subfigure[branch and bound]{
			\includegraphics[width=0.3\linewidth]{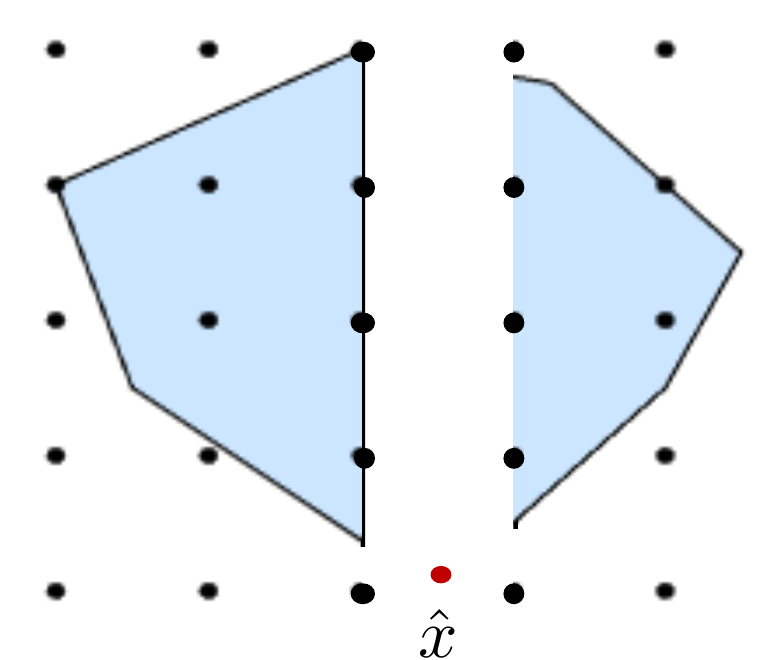}}
		\subfigure[Cutting-plane]{
			\includegraphics[width=0.3\linewidth]{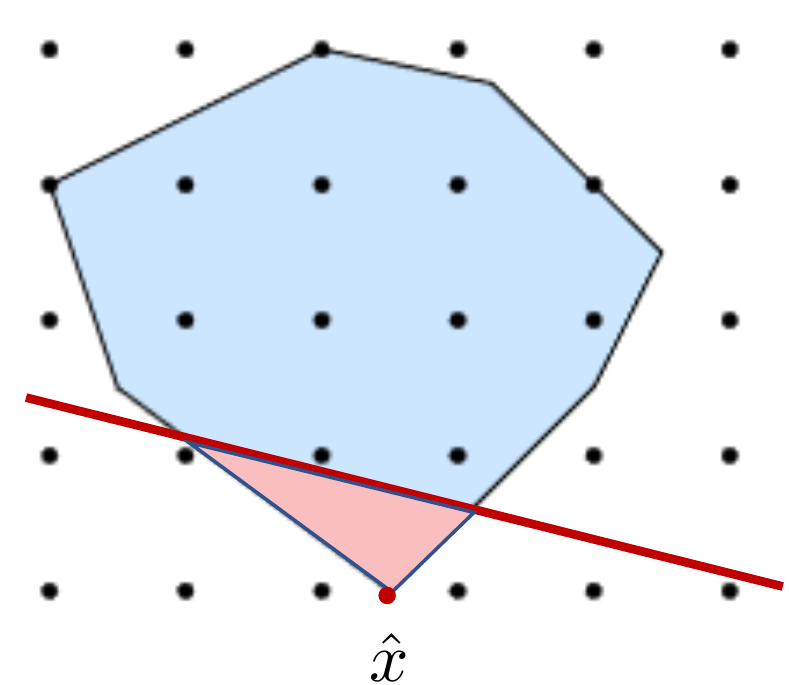}}
	\end{tabular}
	\caption{Tighting LP relaxation by branching and cutting planes.} 
	\label{fig: cutplane}
	\centering
\end{figure}

Cutting planes are additional linear constraints violated by the current LP solution but do not cut off integer feasible solutions. Specifically, cutting plane (sometimes called valid inequalities) methods repeatedly add cuts to the LPs, excluding some part of the feasible region while conserving the integral optimal solution so that the LP relaxation can be tightened. The difference between tightening LP relaxation by branching and cutting planes is illustrated in Figure~\ref{fig: cutplane}.

Depending solely on cutting plane methods is intractable for solving MILPs, and thus, they are always combined with B\&B to tighten the bound further for pruning the tree. According to where cutting planes are generated at the root or subproblems in the B\&B trees, there are two algorithms called \emph{cut-and-branch} and \emph{branch-and-cut}, respectively. The former only generates cutting planes at the root of the B\&B trees, while the latter also produces cutting planes at subproblems. The branch-and-cut is the core of state-of-the-art commercial IP solvers and has been applied in Mitchell~\cite{branchandcut}, Padberg and Rinaldi~\cite{bc2}, and Balas et al.~\cite{bc3}. Moreover, in branch-and-cut, globally valid cuts and locally valid cuts should be distinguished since cuts locally generated at a particular node may be invalid for other nodes, while valid global inequalities can be used for all subproblems. Next, we will briefly introduce cutting planes useful in solving MILPs, including valid inequalities for general integer programs and problems with specific structures.

\subsubsection{General-purpose Cuts}
One of the most fundamental cutting plane methods for MILPs is Gomory's method. \emph{Gomory's fractional cut} was proposed by Gomory~\cite{bams/1183522679} to solve pure integer linear programmings. This method is based on the simplex tableau and has been proved to converge in finite time. From the geometric perspective, Chv\'{a}tal et al.~\cite{V1973Edmonds} illustrated that a convex hull of the integer feasible solutions could be obtained by adding a finite number of Gomory's cuts, known as  Chv\'{a}tal-Gomory cutting-planes. However, Marchand et al.~\cite{marchand2002cutting} have shown that this approach still fails if a problem involves continuous variables. Then, Gomory's extended the fractional cuts to deal with MILPs, known as \emph{Gomory's mixed integer cuts} (GMI) in Gomory~\cite{1960AN}. It has been the first general-purpose cutting planes successfully applied with a branch and cut framework to solve MILPs. Bonam et al.~\cite{bonami2008projected} have shown that for $41$ MILPLIB instances, GMI cuts can help to reduce the integrality gap by $24\%$ on average. In general, Gomory’s mixed integer cut dominates Gomory's fractional cut. Nonetheless, numerical issues in Gomory’s approach prevent pure cutting plane methods from being effective in practice. 

\emph{Mixed integer rounding} (MIR) cuts were introduced by Nemhauser and Wolsey~\cite{1990A}, which are more general than GMI cuts, and gave a complete description for any mixed 0-1 polyhedron. Marchand and Wolsey~\cite{marchand1998polyhedral, marchand2001aggregation} showed that MIR inequalities could produce strong cutting planes for structured mixed integer programs and solve them effectively. Furthermore, it was proved in Nemhauser and Wolsey~\cite{1990A} that MIR cuts are equivalent to \emph{split cuts} proposed by Cook et al.~\cite{1990Chv}. \emph{Lift-and-project} cuts shown in Lov\'{a}sz and Schrijver~\cite{L1991Cones} and Balas et al.~\cite{1993A} lift the LP relaxation into a higher-dimensional space and find valid cuts in this space, and then project valid cuts back into the original space.

\subsubsection{Problem-specific Cuts} 
Apart from the above ``general-purpose" valid inequalities, there are also some methods considering the structure of specific problems and obtaining stronger inequalities. For example, \emph{Flow cover cuts} were introduced by Padberg et al.~\cite{flowcover1985} and generalized in Van Roy and Wolsey~\cite{VANROY1986199}, and Aardal et al.~\cite{aardal1995capacitated} are based on node flow problems, while \emph{clique cuts} proposed by Savelsbergh~\cite{savelsbergh1994preprocessing} are usually applied on conflict graphs, shown in Atamt{\"u}rk et al.~\cite{atamturk2000conflict}. \emph{Knapsack cover cuts} presented in Atamtuerk~\cite{2005Cover} regard constraints as separation knapsack problems, and they are one of the first cutting planes incorporated into commercial MILP solvers~\cite{crowder1983solving}. \emph{Lifted cover inequalities} in van  de  Leensel et al.~\cite{van1999lifting} use lifting to strength cover inequalities and yield a large class of facet-defining inequalities, which have been used successfully in general-purpose branch-and-cut algorithms~\cite{1993A}. 

Selecting which cutting plane to add is a non-trivial task. Modern solvers decompose cutting plane selection into two phases, maintaining valid inequalities in \emph{cut-pool} firstly, then ranking these cuts and selecting some of them to tighten the feasible region. Table 7 in Achterberg and Wunderling~\cite{Achterberg2013} details the contribution of different cutting-plane methods in CPLEX 12.5, indicating that MIR cuts are the most useful, followed by Gomory cuts and knapsack cover cuts, significantly outperforming other cuts. However, general-purpose cutting planes are mostly seen from theoretical interest and analysis. Dey and Molinaro~\cite{2018Theoretical} illustrated several questions that need to be considered in the cutting-plane selection and analyzed existing theoretical challenges in understanding and addressing these issues.

\section{Machine Learning}\label{s3}
\indent \indent In this section, we provide an introduction of the traditional machine learning framework, with the aim to make the reader familiar with some essential concepts which are required to understand the remainder of the survey. 
\subsection{Supervised Learning}
\indent \indent In supervised learning, a set of pairs $\mathcal{D}_n=\{(\mathbf{x}_1,\mathbf{y}_1),(\mathbf{x}_2,\mathbf{y}_2),\ldots,(\mathbf{x}_n,\mathbf{y}_n)\}$ consisting of input (features) $\mathbf{x}_i$ and output (target) $\mathbf{y}_i$ is provided with the aim to find a function $f$ that for every input has a prediction as close as possible to the provided output. The search for  a proper function is based on an optimization problem over a certain (parametrized) family of functions $\{f_\theta \mid \theta \in \mathbb{R}^p \}$ and this procedure is called learning. Together with the family functions, a loss function $l:(f_\theta,\mathcal{D}_n)\rightarrow\mathbb{R}$ is  to compute the discrepancy between the prediction and the target, which is task-dependent. In supervised learning problems, dependent on the target type, two main predictive tasks can be identified  as follows:
\begin{enumerate}
	\item Classification: the output is a qualitative label which can differentiate between $m\in\mathbb{N}$ categories. By encoding the label $y\in\{1,\ldots,m\}$ into a vector $\mathbf{y}\in \mathbb{R}^m$, the accuracy  of a classifier $f$ can be measured by the loss function $l(f,(\mathbf{x,y}))=\mathbf{I}_{\{f(\mathbf{x}) \neq \mathbf{y}\}}$, where $\mathbf{I}_{\{\cdot\}}$ represents  the indicator function. 
	\item Regression: the output is a quantitative value $\mathbf{y}\in \mathbb{R}^m$ and the regressor $f$ outputs the expected value of  $\mathbf{y}$ given $\mathbf{x}$. A commonly used loss function in this setting is the quadratic error $l(f,(\mathbf{x,y}))=||f(\mathbf{x})-\mathbf{y}||^2$.
\end{enumerate} 

Generally, the problem has a statistical nature, i.e., the data ($\mathbf{x_i,y_i}$) is a realization of random variables (${X,Y}$) which follows a joint probability distribution ${P}$. With the goal to find the optimal function as a minimizer of the expected value of $l(f_\theta({X}),{Y})$ under the probability distribution $P$, the supervised learning problem can be formulated as
\begin{equation}\label{e1}
\min\limits_{\theta \in \mathbb{R}^p} \mathbb{E}_{{X,Y}\sim {P}} l(f_\theta({X}),{Y}).
\end{equation}

For example, $\theta$ could be the weights of a linear function and in this case  $f_\theta$ is a linear model. Moreover, since the probability distribution is unknown, instead of minimizing the expected risk, one aims at minimizing the empirical risk by using the examples of finite data $\mathcal{D}_n$  and the optimization problem solved is
\begin{equation}\label{e2}
\min\limits_{\theta \in \mathbb{R}^p} \frac{1}{n}\sum_{i=1}^nl(f_\theta(\mathbf{x}_i),\mathbf{y}_i).
\end{equation}

It is easy for a model to achieve a good performance on the given data. However, one usually hopes the learned model achieve a good performance on unseen data. This is known as generalization, which is a fundamental property in a learned predictor. Owing to the finite number of observations, the learning problem of an optimal function is a delicate task with many traps, one of them is overfitting. 

Generally speaking, overfitting represents the phenomenon of doing well on the given data but not generalizing to the unseen data. This occurs because the optimal function in \eqref{e2} is obtained by fitting the specific dataset $\mathcal{D}_n$, making the training loss underestimates the expected loss. In particular, the more overfitting occurs, the larger the generalization error (the difference between training loss and the expected loss). Since the expected loss remains inaccessible, we can estimate the generalization error through  evaluating the learned function on a separate test set
$\mathcal{D}_{test}$ with
\begin{equation}\label{e3}
\frac{1}{|\mathcal{D}_{test}|}\sum_{\mathbf{(x,y)}\in\mathcal{D}_{test}}l(f_\theta(\mathbf{x}),\mathbf{y}).
\end{equation}
The aim to find a function which does not overfit is at odds with the goal of searching a complex function to capture the characteristics of the data. This is known as the bias-variance trade-off. 

For the purpose of selecting the best among various trained models, a proper model selection procedure divides the dataset into three parts,  to be used in training, validation and test phases. The validation set is used to estimate the generalization error. Based on this estimate, the model selection can be done and the final generalization error of the selected value is obtained by the test set.

\subsection{Unsupervised Learning} \label{s3.2}
\indent \indent In unsupervised learning,  there are no targets for the task one wants to solve, i.e., the prediction is performed without a supervisor who knows the correct answers. The aim of the problem is to learn a function $f$ capturing certain characteristic of the distribution of the observation. Several tasks in this setting are:
\begin{enumerate}
	\item Clustering: similarities within the input space is discovered in the clustering problem. Data could be grouped within a partition of the  whole space and for a new point, it is able to predict its membership to one or several groups. 
	\item Density estimation: the function $f$ is an estimator for the distribution of the input data. Since there is no labeled target, one can maximize the (log-)likelihood of the observation, i.e., their  probability with regard to the underlying distribution $P$. 
	\item Dimensionality reduction: the transformation of  input data, usually from a high-dimensional space into a lower-dimensional space. The purpose of this problem is to identify some essential characteristics of the input data via extraction or selection.   
\end{enumerate}

Since unsupervised learning has received only a little concern on mixed-integer linear programming and its immediate application seems difficult, in this survey, we are not going to discuss it further. The reader is referred to the classical textbooks (e.g.,\cite{bishop2006pattern},\cite{murphy2012machine},\cite{goodfellow2016deep})  on machine learning.

\subsection{Reinforcement Learning}
\begin{figure}[ht]
	\centering
	\includegraphics[width=0.5\textwidth]{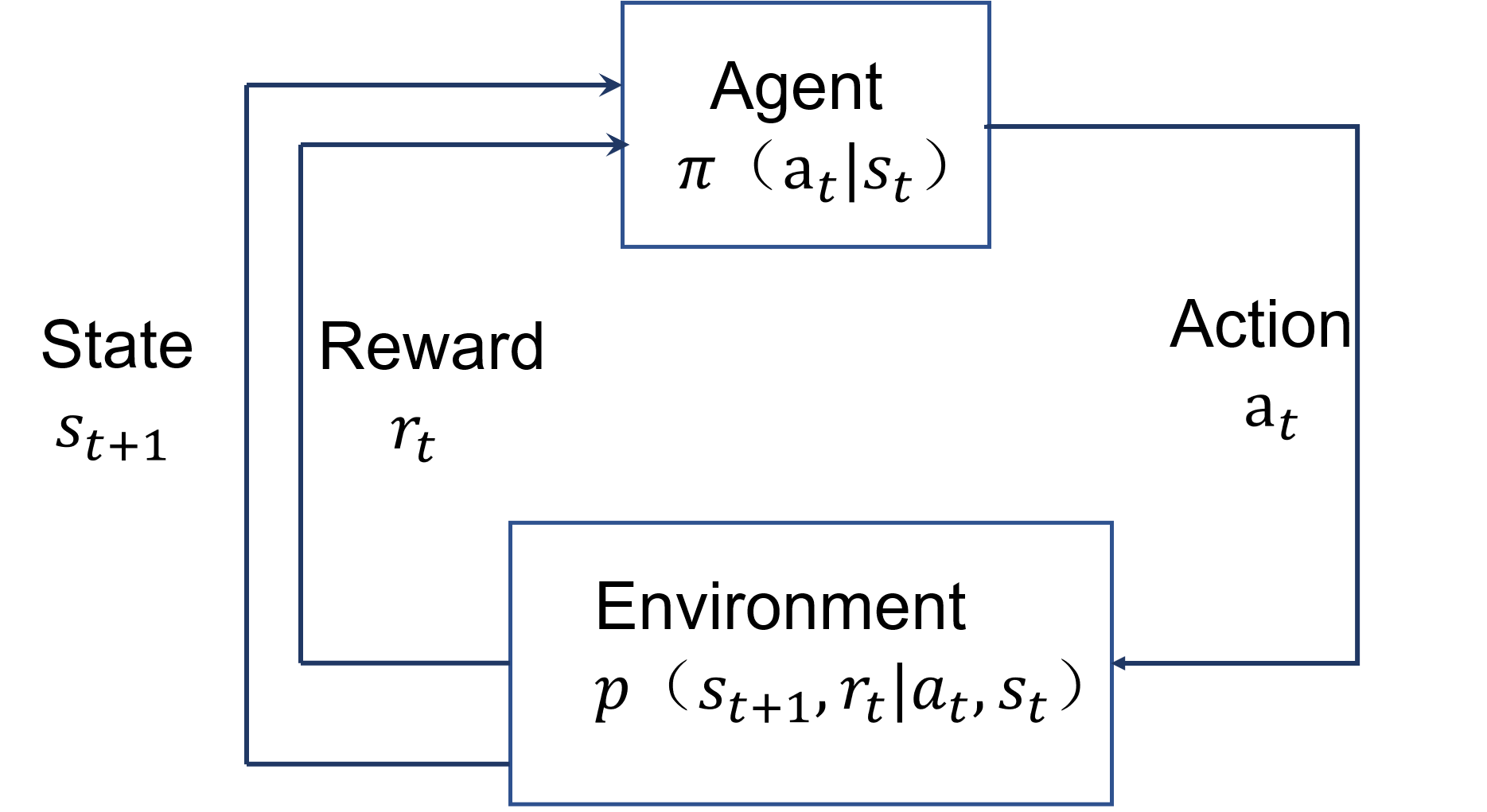}
	\caption{The Markov decision process associated with reinforcement learning.}
	\label{f2}
\end{figure}

\indent \indent In reinforcement learning(RL), an agent interacts with an  environment  for the purpose of maximizing its cumulative reward through trail an error using feedback from its own action. The setting of a Markov decision process(MDP) provides a theoretical framework for reinforcement learning, as illustrated in Fig. \ref{f2}. At each time iteration, the agent is in a given state $s_t\in\mathcal{S}$ of the environment, and subsequently takes an action $a_t\in\mathcal{A}(s_t)$ according to its stochastic policy $\pi(a_t|s_t)$. As a result, the agents enters a new state $s_{t+1}$ with probability $P^{a_t}_{s_t,s_{t+1}}$ and it receives an immediate  reward  $R^{a_t}_{s_t,s_{t+1}}$ from the environment. The aim of reinforcement learning is to learn an optimal policy $\pi: \mathcal{S}\rightarrow \mathcal{A}$ that maximizes the expected discount sum of future rewards (discounted return):
\begin{equation}
G(s_0,\pi)=\mathbb{E}[R^{a_0}_{s_0,s_{1}}+\gamma R^{a_1}_{s_1,s_{2}}+\cdots+\gamma^tR^{a_t}_{s_t,s_{t+1}}+\cdots],
\end{equation}
where $\gamma\in[0,1]$ denotes the discount rate, which is used to model the fact that the future reward is less worth than an immediate one. Given a policy $\pi$ and state $s_t$ (resp. state and action pair $(s_t, a_t)$), the expected return is referred to the value function (resp. state action value function). The value function follows the Bellman equation, thus the learning problem can be formulated as a dynamic programming, and solved approximately.  

A major concern in reinforcement learning is the exploration \textit{vs} exploitation dilemma: choosing between exploring  new states by trying new actions in order to refine the knowledge of the environment for possible improvements in the long term, or exploiting experienced actions which yield a high reward. Moreover, defining a reward is sometimes a difficult task. Reinforcement learning is able to credit the states/actions, bringing about future rewards due to its dynamic programming process. However, it is still challenging since no learning opportunity is  offered until the agent solves the problem. In addition,   it is not guaranteed that the learning converges to the global optimum when the policy is approximated. For more details, the interested reader can refer to \cite{sutton2018reinforcement}, which is an extensive textbook on reinforcement learning. 

\section{Learning-Based Branch and Bound Algorithms}\label{s4}
In this section, we survey the learning techniques to deal with the four key components in B\&B algorithms for MILP. The contributions and limitations of different studies are summarized in this section. In addition, we make some comparisons of different learning methods in these four key components.

\subsection{Learn to Branch}
With the development of learning, alleviating the computational burden in some traditional branching strategies with learning has been a hot topic. Some works have used imitation learning to approximate decisions by observing the demonstrations shown by an expert, while others capitalize on reinforcement learning to explore better branching policies.

\subsubsection{Supervised Learning in Branching}
Since strong branching is the most efficient expert in terms of the number of expanded nodes till now, and its main disadvantage is prohibitive computational cost, most of the works learn to mimic strong branching and use learning to estimate scores quickly. This method is called imitation learning, which can be regarded as supervised learning for decision-making. Some early attempts making use of this idea utilize traditional machine learning methods. Alvarez et al.~\cite{Alvarez14asupervised, doi:10.1287/ijoc.2016.0723, 2016Online} trained a regression model to approximate strong branching scores efficiently. Alvarez et al.~\cite{Alvarez14asupervised, doi:10.1287/ijoc.2016.0723} learned the SB score function by ExtraTree shown in Geurts et al.~\cite{geurts2006extremely}, while~\cite{2016Online} learned by online linear regression. The method proposed in ~\cite{Alvarez14asupervised, doi:10.1287/ijoc.2016.0723} consists of two phases to learn and solve MILPs. In the first phase, the SB decisions are recorded by solving a set of randomly generated problems optimally, and a regressor is learned to predict strong branching scores. Then in the second phase, the learned function is employed for instances from MIPLIB, which is the conventional evaluation benchmark for MILP methods. On the other hand, they presented some significant features to describe the current problem, which are not only complete and precise but also efficient to compute. These features are divided into three types. Specifically, \emph{static problem features} are computed solely from parameters of the original problem, $c, A$ and $b$, thus calculated once and for all. They gave an overall description of the problem and represent the static state of the problem. \emph{Dynamic problem features} describe the state of a specific candidate concerning the LP solution at the current node. Besides, \emph{dynamic optimization features} extract the overall statistical effect of the candidate about the optimization process. These features mainly take into account the variables' roles at the current nodes. Zarpellon et al.~\cite{zarpellon2020parameterizing} generalized features in the space of B\&B search trees, leading to a more flexible and adaptive variable selection. Therefore, feature design is a crucial work in learning-based methods. Alvarez et al.~\cite{doi:10.1287/ijoc.2016.0723} revealed three desirable properties of branching features, followed by many later works. First, features should be \emph{size-independence}, which is intuitive and fundamental to solve large-scale MILPs. Second, features should be \emph{invariant} to some irrelevant changes, like permutating rows or columns of matrix $A$. Experiment results show that the learned branching strategy compares favorably with SB, but its performance is slightly below that of RB. When solving MILPs, the learned strategy performs badly only on a small number of instances, but it is faster than RB in $11/30$ cases and faster than FSB in $21/30$ instances.
Inspired by the idea behind RB, Alvarez et al.~\cite{2016Online} made use of online learning to imitate SB. For a candidate, if its SB score has been computed a certain number of times, it would be deemed reliable, and its SB score could be approximated by learning. Otherwise, feature vectors and scores of unreliable candidates can be put into the training set. Hence, in contrast to~\cite{Alvarez14asupervised, doi:10.1287/ijoc.2016.0723}, the training data was generated during the B\&B process on-the-fly, and thus, no preliminary phase in needed.
Khalil et al.~\cite{Khalil_2016} learned string branching scores in an on-the-fly fashion, without an upfront offline training phase on a large set of instances, so no preliminary phase is required to record expert behavior, and computing time is  saved. On the other hand, this method is instance-specific since it learns from the expert at the beginning of the tree, and the learned ranking function is then used for branching seamlessly. Ranking formulation with binary labels in this work is natural for variable selection, as the score itself is not important and what we care about is ``whether" this candidate is effective. Moreover, binary labels relax the definition of ``best" branching variables, allowing us to consider many good variables that also have high scores in the learning. They avoid learning from correctly rank candidates with low scores, which further saves time. Experiment results show that imitating SB with machine learning outperforms SB and PC in terms of the solved instances and the number of nodes, and even solves more instances than the CPLEX-D solver. Although the running time of the learned policy is more than pseudocost branching for easy instances, machine learning runs fastest when solving medium and hard MILPs. These early works reveal the potential of taking advantage of machine learning to speed up large-scale MILPs. 

Moreover, to tackle tedious parameters tuning, Balcan et al.~\cite{pmlr-v80-balcan18a} learned the optimal parameter setting for the instance distribution. A certain application domain can be modeled as a distribution over problem instances, and samples can be accessed by the algorithm designer so that a nearly optimal branching strategy can be learned. Here, the optimal branching policy means the optimal convex combination of different branching rules. This work theoretically proved that using a data-independent discretization of the parameters to find an empirically optimal B\&B configuration is impossible, indicating its intrinsic adaptiveness. In other words, the effect of branching parameters on the average number of expanded nodes varies significantly with the applications.

\begin{figure}[!hbt]
	\centering
	\includegraphics[width=0.8\textwidth]{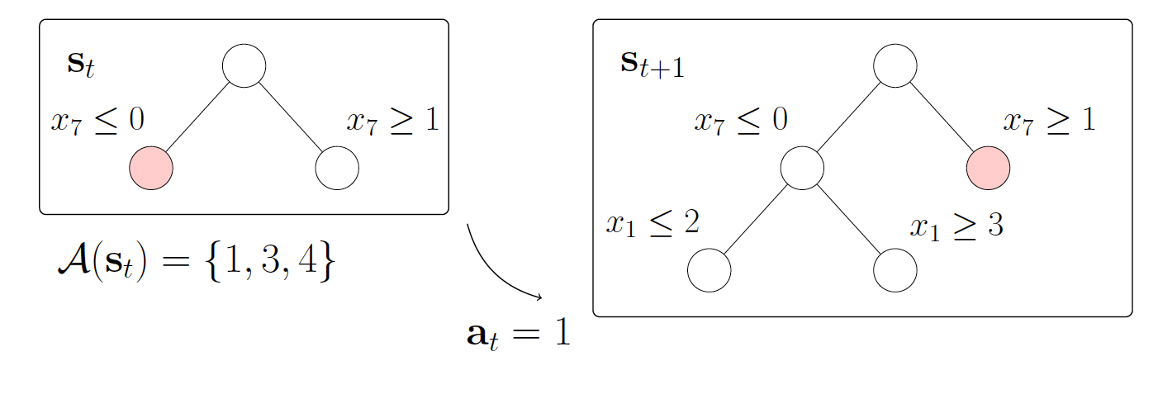}
	\caption{Variable selection in B\&B integer programming algorithm as an MDP. In the left figure, a state $s_t$ includes the B\&B
		tree with a leaf node chosen by the solver to be expanded next (in pink).
		In the right figure, a new state $s_{t+1}$ is obtained by branching over the variable $a_t = x_1$ \cite{gasse2019exact}.}
	\label{gnn_mdp}
\end{figure}

\begin{figure}[!hbt]
	\includegraphics[width=0.9\textwidth]{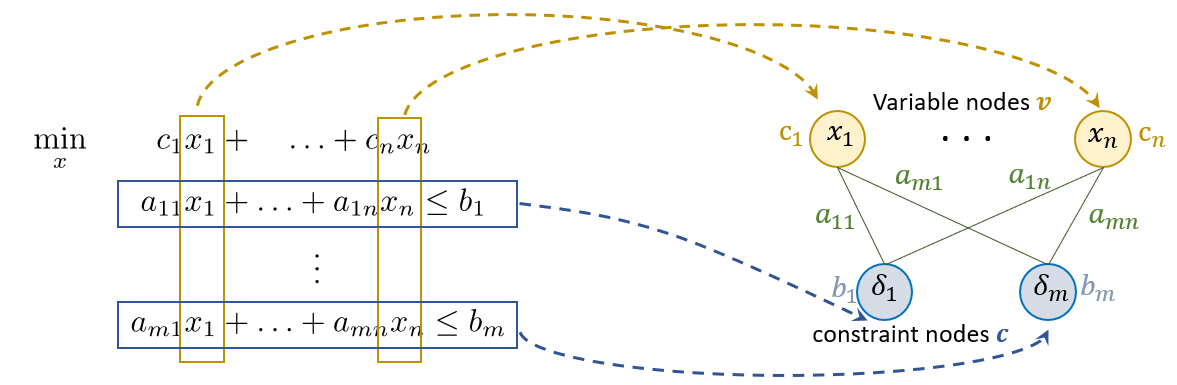}
	\caption{Bipartite graph presentation of MILP \cite{gasse2019exact}. The bipartite graph consists of two sets of nodes, the variable sets $\{x_1,\ldots,x_n\}$ and the constraint nodes $\{\delta_1,\ldots,\delta_m\}$. The edge connecting the variable node and the constraint node denotes the coefficients of the MILP.} 
	\label{bigraph}
\end{figure}

\begin{figure}[!hbt]
	\includegraphics[width=0.9\textwidth]{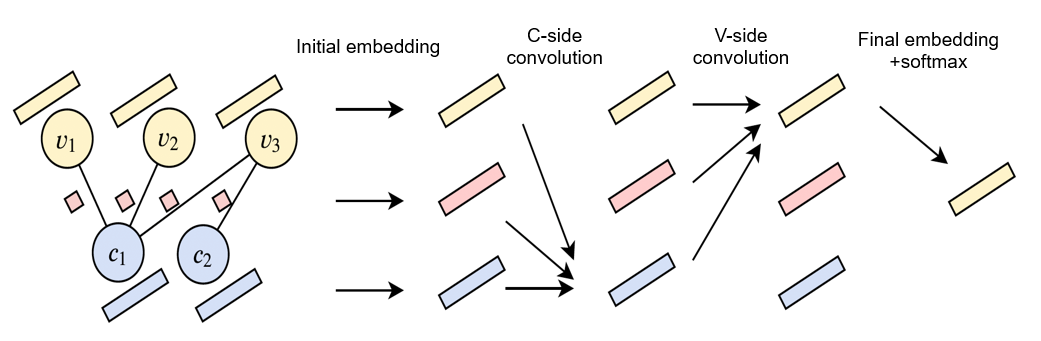}
	\caption{The architecture of the GCNN model in \cite{gasse2019exact}.} 
	\label{gnn}
\end{figure}

Nowadays, deep learning has achieved huge success in various fields, with the power to process huge amounts of data. Many works have made use of powerful neural networks to speed up variable selections. Due to the bipartite graph representing of mixed integer linear programming (Fig. \ref{bigraph}), it is natural to encode the branching policies to a graph convolutional neural network (GCNN), which speeds up the MILP solver by reducing the amount of manual feature engineering. Moreover, the previous statistical learning of branching strategies is only able to generalize to similar instances while the GNN model has a better generalization ability since it can model problems of arbitrary size. Gasse et al. \cite{gasse2019exact} first adopted imitation learning and a dedicated graph convolutional neural network model to address the B\&B variable selection problem. In this work, the problem was formulated by the task of searching the optimal policy of an MDP, as illustrated in Fig. \ref{gnn_mdp}. Since the graph structure is the same for all LP relaxation in the B\&B tree represented as a bipartite graph with node and edge feature, the cost of extraction is reduced to a great extent. By adopting imitation learning, a GCNN model was trained to approximate the strong branching policy, a very good but computationally expensive expert policy. The overview of their architecture is depicted  in Fig. \ref{gnn}. The bipartite representation is taken as input of the model and a single graph revolution is performed in the form of two half convolutions. A probability distribution over the candidate branching strategies is finally obtained by discarding the constraints node and adopting a multi-layer preceptron to the variable nodes, combined with a softmax activation.  Their resulting branching policy is shown to  perform better than previously proposed methods for branching on several MILP problem benchmarks and generalize to larger instances than trained on. In order to  make the GNN model more competitive on CPU-only machines, Gupta et al. \cite{gupta2020hybrid}  devised a hybrid branching model that uses a GNN model only at the initial decision point and a weak but fast predictor, such as multi-layer perceptron, for subsequent steps. The proposed hybrid architecture improves the weak model by extracting high-level structural information at the initial point by the GNN model and preserves the original GNN model's ability to generalize to harder problems than trained on. Nair et al. \cite{nair2020solving} adopted imitation learning to obtain a MILP brancher, where the GNN approach was expanded by implementing a large amount of parallel GPU computation. By mimicking an ADMM-based expert and combining the branching rule and primal heuristic, their work is advantageous over the  SCIP \cite{gamrath2020scip} in terms of solving time on five benchmarks in real life.

However, most of the above works specialize the branching rule to distinct classes of problems, obtaining generalization ability to larger but similar MILP instances \cite{Alvarez14asupervised, doi:10.1287/ijoc.2016.0723, 2016Online, Khalil_2016}. Besides, these works lack a mathematical understanding of branching due to its inherent exponential nature. With the hypothesis that parameterizing the underlying space of B\&B search trees can aid generalization to heterogeneous MILPs, Zarpellon et al.~\cite{zarpellon2020parameterizing} introduced new features and architectures to represent branching, making the branching criteria adapt to the tree evolution and generalizing across problems not belonging to the same combinatorial class. Specifically, to capture the dynamics of the B\&B process linked to branching decisions, some features were proposed to describe the roles of candidates in the B\&B process. At a certain branching step, besides some features capturing statistics and multiple roles of a variable throughout the search, denoted as $C_t$, there are some features, $Tree_{t}$, describing the state of the search tree, such as the growth rate, the composition of the tree, and the bound quality of the current node. All these features, $\{C_{t}, Tree_{t}\}$, are not explicitly dependent on the specific problem parameters and evolve with the search, different from static and parameters-dependent features utilized by the above works. This work first used a baseline DNN (NoTree) that only uses $C_t$ as inputs to demonstrate the significance of search-based features. The output layer equipped with the softmax produces a probability distribution over the candidate set, indicating the probability for each variable to be selected. Then, $Tree_{t}$ was input to the NoTree by gating layers, called as TreeGate models. Considering that people rarely use SB in practice, this work chose the SCIP default branching scheme, relpscost~\cite{GleixnerBastubbeEifleretal.2018}, a reliability version of HB, as a more realistic expert. In the experiment, $27$ heterogeneous MILP problems were partitioned into $19$ train and $8$ test problems. Both NoTree and TreeGate outperform GCNN, RB, and PC in terms of the total number of nodes explored. Since GCNN struggles to generalize across heterogeneous instances, it expands around three times as many nodes as this method does. On the other hand, the test accuracy of TreeGate test accuracy is $83.70\%$, improving $19\%$ over the NoTree model, while the accuracy of GCNN is only $15.28\%$. The comparison between the above papers are shown in Table \ref{vs1} and \ref{vs2}.

\begin{table}[]
	\scriptsize
	\caption{Machine Learning-Based Variable Selection Comparison}
	\label{vs1}
	\begin{tabular}[c c c c]{p{0.1\textwidth}<{\centering}p{0.235\textwidth}<{\centering}p{0.235\textwidth}<{\centering}p{0.235\textwidth}<{\centering}}
		\toprule\toprule
		& Alvarez et al., 2016 \cite{doi:10.1287/ijoc.2016.0723}   & Alvarez et al., 2016  \cite{2016Online}                     & Khalil et al., 2016 \cite{Khalil_2016}  \\ \midrule
		Learning approach            & Offline extraTrees for regression                                                                                                                                                                                                                              & Online linear regression                                                                                                                                 & SVM with rank formulation                                                                                                                                                        \\ \midrule
		Features                     & \begin{tabular}[c]{p{0.235\textwidth}}Static problem features\\ Dynamic problem features\\ Dynamic optimization features\end{tabular}                                                                                                                                     & Features from Alvarez et al. \cite{doi:10.1287/ijoc.2016.0723}                                                                     & \begin{tabular}[c]{p{0.235\textwidth}}72 atomic features computed on the node LP and candidate variable; \\ Interation features computed from a product of two atomic features\end{tabular} \\ \midrule
		Expert                       & SB                                                                                                                                                                                                                                                             & SB                                                                                                                                                       & SB                                                                                                                                                                               \\ \midrule
		Compared algorithm(s)        & \begin{tabular}[c]{p{0.235\textwidth}}Random branching\\    MIB\\    NCB (nonchimerical branching)\\    FSB\\    RB\end{tabular}                                                                                                           & \begin{tabular}[c]{p{0.235\textwidth}}FSB\\    RB\\    Offline learned branching\end{tabular} & \begin{tabular}[c]{p{0.235\textwidth}}CPLEX 12.6.1 default\\    SB\\    PC\\    SB + PC\end{tabular}                                                                           \\ \midrule
		Measure(s)                    & \begin{tabular}[c]{p{0.235\textwidth}}closed gap\\    \# solved problems\\    \# nodes\\    time\end{tabular}                                                                                                                                                             & \begin{tabular}[c]{p{0.235\textwidth}}\#nodes\\    time\end{tabular}                                                                                                & \begin{tabular}[c]{p{0.235\textwidth}}\# unsloved problems\\    \# nodes\\    total time\end{tabular}                                                                                       \\ \midrule
		Advantage(s)                   & \begin{tabular}[c]{p{0.235\textwidth}}Reduce time required to take a decision, and thus close the gap by exploring more  nodes;\\    
			Run fastest when cuts and heuristics are used by CPLEX.\end{tabular}                                                  & Save time in an online fashion.                                                                                                                          & \begin{tabular}[c]{p{0.235\textwidth}}Allow one to take into account many good candidates in the learning;\\    Perform best in medium and hard problems.\end{tabular}                      \\ \midrule
		\multirow{2}{*}{Limitation(s)} & \begin{tabular}[c]{p{0.235\textwidth}}Perform worse than RB under the setting     of node limit and the setting of no cuts and heuristics;
			\\    Cannot generalize to very large or    heterogenous problems;
			\\    Require expensive training phase.\end{tabular} & Cannot generalize to heterogenous  problems.                                                             & Focus on static and parameters-dependent properties of problems.                                                                                                                 \\ \cline{2-4} 
		& \multicolumn{3}{c}{Be limited by the expert.}                                                 \\ \bottomrule	
	\end{tabular}
\end{table}
\begin{table}[]
	\scriptsize
	\caption{Deep Learning-Based Variable Selection Comparison}
	\label{vs2}
	\begin{tabular}[c c c c c]{p{0.1\textwidth}<{\centering}p{0.19\textwidth}<{\centering}p{0.19\textwidth}<{\centering}p{0.19\textwidth}<{\centering}p{0.19\textwidth}<{\centering}}
		\toprule\toprule
		& Gasse et al., 2019 \cite{gasse2019exact}               & Gupta et al., 2020  \cite{gupta2020hybrid}                   & Nair et al., 2020 \cite{nair2020solving}                     & Zarpellon et al., 2020 \cite{zarpellon2020parameterizing}     \\ \midrule
		Learning approach  & GCNN  & HyperGNN-FiLM                 & GCNN & NN with feature gating \\ \midrule
		Features                     & \begin{tabular}[c]{p{0.19\textwidth}}5 features for the constraint;\\    13 features for the variable; \\    1 feature for the edge.\end{tabular}                                                    & \begin{tabular}[c]{p{0.19\textwidth}}Root node: 19 features from Gasse et al. \cite{gasse2019exact}\\    Remaining node: 72 features from Khalil et al. \cite{Khalil_2016}\end{tabular}             & Features from  Gasse et al. \cite{gasse2019exact}                 & \begin{tabular}[c]{p{0.19\textwidth}}25 features representing set of candidate\\    variables\\    61 features encoding dynamic state of tree\\    search\end{tabular} \\ \midrule
		Expert  & SB& SB   & ADMM-based FSB  & SCIP default \\ \midrule
		Compared algorithm(s)        & \begin{tabular}[c]{p{0.19\textwidth}}RPB (reliability pseudocost)\\    FSB\\    Offline learned branching\\    SB + ML\\    LMART  \cite{hansknecht2018cuts}\end{tabular}            & \begin{tabular}[c]{p{0.19\textwidth}}RPB\\    PB(heuristic Pesudocost)\\    SB + ML\\    ExtraTree\\    GCNN\end{tabular}                                                                                                             & \begin{tabular}[c]{p{0.19\textwidth}}FSB\\    SCIP 7.0.1 default\end{tabular}                                                                                                              & \begin{tabular}[c]{p{0.19\textwidth}}Random branching\\    PC\\    SCIP default\\    GCNN\end{tabular}                                                                    \\ \midrule
		Measure(s)                    & \begin{tabular}[c]{p{0.19\textwidth}}\# nodes\\    time\end{tabular}                                                                                                                                 & \begin{tabular}[c]{p{0.19\textwidth}}\# nodes\\    time\end{tabular}                                                                                                                                                                  & time                                                                                                                                                                             & \# nodes                                                                                                                                                     \\ \midrule
		Advantage(s)                   & \begin{tabular}[c]{p{0.19\textwidth}}Reduce feature calculation cost; \\    Faster solving time than the solver default procedure;\\    Generalize to larger instances than trained on.\end{tabular} & Show improvements on CPU-only hardware.                                                                                                                                                                                     & Improve SCIP  in solving time over real-life benchmarks and MIPLIB.                                                                                                              & Generalize across heterogenous MILPs; Handle candidates of varying size.                                                                                     \\ \midrule
		\multirow{2}{*}{Limitation(s)} & \begin{tabular}[c]{p{0.19\textwidth}}Uncompetitive on CPU-only machines;\\    Tailored to the type of MILP problems it is trained on.\end{tabular}                                                   & \begin{tabular}[c]{p{0.19\textwidth}}Only evaluate performance of the trained model on small instances;\\    A poor generalization capability for Maximum Independent Set problem;\\    Focus on specific problem types.\end{tabular} & \begin{tabular}[c]{p{0.19\textwidth}}Sometimes fail to find an optimal or near-optimal solution;\\    Only generalize to unseen instances from the same problem distribution.\end{tabular} &                                                                                                                                                              \\ \cline{2-5} 
		& \multicolumn{4}{c}{Be limited by the experts.}                                  \\ \bottomrule
	\end{tabular}
\end{table}

\subsubsection{Reinforcement Learning in Branching}
One of the main limitations of imitation learning is that the expert bounds the performance of the learned strategy. However, expert knowledge is not satisfactory sometimes. Sun et al.~\cite{sun2020improving} analyzed why imitating SB for learning branching strategy is not a wise choice: SB yielding small B\&B trees is a consequence of reductions from solving branch LP instead of its decision quality, and these reductions cannot be learned by imitating a branching policy. They designed experiments that eliminate the side-effect of reduction obtained in solving branch LP and found that SB has poor performance. Thus, researchers hope to propose better methods of selecting variables.

Sun et al. ~\cite{sun2020improving} designed a novel primal-dual policy network over reduced LP relaxation and a novel set representation for the branching strategy. To encourage the agent to solve problems with as few steps as possible, the reward is set to $r_t = 1$ at each time step. Although the primal-dual policy is similar to GCN, it uses a dynamic reduced graph by removing redundant fixed variables and trivial constraints during the process, which gives a more precise and accurate description of the state. On the other hand, instead of edge embedding in GCN, primal-dual policy relies on simple matrix multiplication, which further saves computational time. Here, set representation and optimal transport distance (or Wasserstein distance) are used to define the novelty score. By regarding every subproblem $Q$ generated from an instance by a branch policy $\pi$ as a polytope, the collection of its subproblems can be defined as $b(\pi, Q) = \{R_1, \dots, R_H\}$. For each subproblem, $R_i$, the weight function $w(\cdot)$ represents the number of feasible point in the associated polytope. The distance between two subproblems $R_i$ and $R_j$ are defined as $d(R_i, R_j):=\| g_i-g_j \|_{1}$, where $g_i$ and $g_j$ are their mass centers. Then the optimal transport distance between two policy representations is computed as
\begin{equation}
D(b_1, b_2)=\min_{\Gamma}\sum_{i,j} \Gamma_{ij}W_{ij}(b_1, b_2), \quad 
\text{s.t.} \quad \Gamma\bm{1}=p(b_1), \Gamma^{T}\bm{1} = p(b_2),
\end{equation}
where $p(b) \in \Delta^{H-1}$ is a simplex mapped from b by normalizing the weights. Therefore, given a collection of older policies $M$ and an instance $Q$, novelty score is defined as
\begin{equation}
N(\theta, Q, M) = \frac{1}{k}\sum_{\pi_{j} \in k\text{NN}(M, \theta)} D(b(\pi_{\theta}, Q), b(\pi_{j}, Q)),
\end{equation}
where $k$NN is the $k$ nearest neighbor of $\pi_{\theta}$ in $M$. Equipped with this novelty score, Novelty Search Evolution Strategy was proposed to encourage exploration in reinforcement learning. Their experiments compared the RL agent with SVM and GCN, two competitive learning methods in imitation learning, in addition to three SCIP's branching rules. The RL agent performs best in terms of the running time, the number of expanded nodes, and primal bounds. However, about obtaining a higher dual value, the agent performs worst initially, but it finally obtains the best result, indicating its non-myopic policy. Different from~\cite{RLVB1}, this work can transfer to larger instances, and the performance of the RL agent is significantly superior to FSB, RPB (reliability pseudocost branch), SVM, and GCN.

Although reinforcement learning can potentially produce a better branching strategy than the expert, the agent performance is limited by the episode length, and it learns inefficiently at the beginning due to little experience.

\subsubsection{Dynamic Approach for Switching Branching Heuristics}
Based on the observation of the highly dynamic and sequential nature of B\&B, Di Liberto et al.~\cite{di2016dash} believed that there is no single branching heuristics given in Section~\ref{BVS} that would perform the best on all problems, even on different subproblems induced from the same MILP. Thus, the efficiency of the search can be much improved if we adopt the correct branching method at the right time during the B\&B search. Motivated by portfolio algorithms that attempt to predict the best heuristic for a given instance, this work proposed an algorithm named Dynamic Approach for Switching Heuristics (DASH). Based on the defined features, a clustering of problems was learned with the g-means algorithm at the first step. Then the correct assignment of branching methods to clusters is learned during an offline training phase with a similar method shown in Kadioglu et al.~\cite{kadioglu2012non}. With the search depth increasing, the instance tends to shift to a different cluster. When such a change occurs, the heuristic would be switched to a new one that best fits the current cluster. Numerical results show that DASH outperforms static and randomly switching heuristics methods in terms of the running time, indicating the benefit of dynamics and adaptiveness of the switching methods. Nevertheless, Lodi and Zarpellon~\cite{lodi2017learning} observed that the upfront offline clustering conflicts with the ever-changing characteristic of the tree evolution and somehow affects time efficiency. 

\subsection{Learn to Select}
As shown in \cite{linderoth1999computational}, the effectiveness of different selection methods depends on the problem type. It is preferable to seek a selection strategy that adapts to different problem types. We divide the existing learning-based literature into the following two categories, i.e., one is adaption in evaluation criteria, and the other is learning in heuristics.

\subsubsection{Adaption in Evaluation Criteria} \label{s4.2.1}
As summarized in Section \ref{s2.3.3}, the evaluation criteria of a node during a B\&B run are from the following two sides, the feasibility side and the optimality side. If the selection criteria pay more attention to the feasibility side, the selection strategy will perform more like the depth-first search strategy, while on the other side, the selection strategy would perform more like the best-first strategy.

Borrowing a RL vocabulary, some grade of adaption could be pursued in the combination of best-first and depth-first strategies to balance exploration and exploitation in the B\&B run. Sabharwal et al. \cite{sabharwal2012guiding} exploited RL framework. The score of a node $ N $ is a weighted sum of two terms
\begin{equation}
\text{score}(N)=\text{estimate}(N)+\varGamma\dfrac{\text{visits}(P)}{100\text{visits}(N)},
\end{equation}
where estimate$( N )$ is some measure of the quality of node $ N $, $ P $ is the parent node of $ N $ and visits$ (\cdot) $ counts the number of times the search algorithm has visited a node. The parameter $ \varGamma $ balances the tradeoff between exploitation(first term) and exploration(second term), and nodes with a higher estimate or have been visited less time than their siblings will be pursued first. This geometric means of runtime, the number of searched nodes, and the simplex iterations are improved compared with three other selection strategies: best-first, breath-first, and CPLEX default heuristic. The improvement is gained due to a balanced usage of best-first and breath-search-like schemes. However, this work treats every node of a B\&B tree equivalently, ignoring the uniqueness of each node's feature during the B\&B run. An interesting question arises about employing the node feature into the scoring system to improve the learning quality.

\subsubsection{Learning in Heuristics in Selecting}\label{s4.2.2}
The main idea of learning in heuristics is rating each node based on a weighted sum of criteria and choosing the node with the highest rating. The first proposed use of learning methods within a heuristic search procedure comes from Glover, and Greenberg \cite{glover1986future,glover1989new} which adjusts the wights offline using a learning procedure. Ans\'{o}tegui et al. \cite{ansotegui2017reactive} induced the hyper configurable reactive search recently to learn the parameters of a metaheuristic online with a linear regression, where the regression weights are tuned offline with the GGA++ algorithm configuration \cite{ansotegui2015model}. 

Daum\'{e} et al. \cite{daume2009search} and Chang et al. \cite{chang2015learning} converted the solving problem into a sequential decision-making problem, for which a policy is then learned or improved. A fundamental limitation of their work is that there is no correction due to the greedy search at each test time; thus, the obtained solution sequence will have deviations from the optimal one.

He et al. \cite{he2014learning} instead proposed a method to learn the node selection strategy in a B\&B run by imitation learning. They categorized their features into three groups:
\begin{enumerate}
	\item Node features include bounds, objective function estimation at a given node, indications about the current depth, and the (parental) relationship for the past proceeded node.
	\item Branching features describe the variable whose branching led to a given node, including pseudocosts, variable's value modifications, and bound improvement.
	\item Tree features consider measures such as the number of solutions found, global bounds, and gaps (see in Definition \ref{d3}).
\end{enumerate}
They assumed that a small set of solved problems are given at training time and the problems to be solved at the test time are of the same type. The node selection policy is learned to repeatedly pick a node from the queue of all unexplored nodes that mimics a simple oracle that knows the optimal solution in advance and only expands nodes containing the optimal solution. This learning-based selection method finds solutions with a better objective and establishes a smaller gap, using less time demonstrated on multiple datasets compared to SCIP. However, the author leaves the seek of certified optimality for speed as future work. In addition, this work is problem-dependent which requires the test data to have the same type of training data. Yilmaz and Smith \cite{yilmaz2021study} used a similar method in \cite{he2014learning} to select the direct children at non-leaf nodes. 

Hottung et al. \cite{hottung2020deep} developed a new method that integrates deep neural networks (DNNs) into a heuristic tree search to decide which branch to choose next, namely, the Deep Learning Heuristic Tree Search (DLTS). DLTS is able to achieve a high level of performance with no problem-specific information. The problem-specific information is almost exclusively provided as input to the DNN, where the DNNs are trained offline via supervised learning on existing (near-) optimal solutions. It is shown in \cite{hottung2020deep} that DLTS significantly finds smaller gaps to optimality on real-world-sized instances compared to that for state-of-the-art metaheuristics proposed by Karapetyan et al. \cite{karapetyan2017markov}. In addition, it does not require extra training data since a learned model to fully control decisions is learned during the search. However, as a coin has two sides, the system performance relies on the quality of the provided solutions. In addition, the policy is not adjusted in terms of the runtime and solution quality.

The comparison between the above papers is shown in Table \ref{ns1}.


\begin{table}[!t]
	\caption{Learning-Based Node Selection Comparison}
	\label{ns1}
	\scriptsize
	\begin{tabular}[t t t t t ]{p{0.17\textwidth}<{\centering}p{0.17\textwidth}<{\centering}p{0.17\textwidth}<{\centering}p{0.17\textwidth}<{\centering}p{0.17\textwidth}<{\centering}}
		\toprule
		\toprule
		&
		Sabharwal et al., 2012 \cite{sabharwal2012guiding} &
		He et al., 2014 \cite{he2014learning} &
		Yilmaz and Smith, 2021 \cite{yilmaz2021study} &
		Hottung et al., 2020 \cite{hottung2020deep}\\ \toprule
		Learning approach &
		Reinforcement learning &
		Imitation Learning &
		Imitation Learning &
		Supervised Learning \\\midrule
		Input(s) &
		\begin{tabular}[c]{p{0.17\textwidth}}Normalized LP objective value of   a leaf node\\     \# times that a node  has been   visited\end{tabular} &
		\begin{tabular}[c]{p{0.17\textwidth}}Node features\\      Branching features\\     Tree features\end{tabular} &
		\begin{tabular}[c]{p{0.17\textwidth}}Node features\\      Branching features\\      Tree features\end{tabular} &
		\begin{tabular}[c]{p{0.17\textwidth}}Node features\\      Branching features\\      Tree features\end{tabular} \\\midrule
		Expert &
		&
		Oracle &
		Best-estimate with plunging   (SCIP default) &
		Existing solutions through   search \\\midrule
		Output(s) &  A parameter $ \varGamma $ to   balance the tradeoff between exploitation and exploration &
		\begin{tabular}[c]{p{0.17\textwidth}} The next node to be explored   picked from the queue of unexplored node\\     Top-best pruning action\end{tabular} &
		\begin{tabular}[c]{p{0.17\textwidth}}The direct children at non-leaf   nodes to be  selected\\    k-best pruning action\end{tabular} &
		\begin{tabular}[c]{p{0.17\textwidth}}Trained deep neural   networks\\      Probability to select this branch\\      Estimated lower bound of each node\end{tabular} \\ \midrule
		Compared algorithm(s) &
		\begin{tabular}[c]{p{0.17\textwidth}}Best-first search\\      Breath-first search\\      Depth-first search (DFS)\\      CPLEX default heuristic\end{tabular} &
		\begin{tabular}[c]{p{0.17\textwidth}}Selection along with   pruning\\      Pruning only\\      SCIP with a node limit\\      Gurobi with a time limit\end{tabular} &
		\begin{tabular}[c]{p{0.17\textwidth}}SCIP default\\      DFS\\      RestartDFS\\      He et al., 2014 \cite{he2014learning}\end{tabular} &
		Metaheuristics (Karapetyan et   al., 2017 \cite{karapetyan2017markov}) \\ \midrule
		Measure(s) &
		\begin{tabular}[c]{p{0.17\textwidth}}Geometric means of runtime\\      \# searched nodes\\      \# simplex iterations\end{tabular} &
		\begin{tabular}[c]{p{0.17\textwidth}}Speedup w.r.t. SCIP   default\\      Optimality gap\\      Integrality gap\end{tabular} &
		\begin{tabular}[c]{p{0.17\textwidth}}Geometric means of runtime\\      Optimality gap\end{tabular} &
		\begin{tabular}[c]{p{0.17\textwidth}}Geometric means of runtime\\      Optimality gap\end{tabular} \\ \midrule
		Advantage(s) &
		Improve all the above measures. &
		Find solutions with a better  objective and establish a smaller gap, using less time. &
		\begin{tabular}[c]{p{0.17\textwidth}}Is effective when the ML model  is able to meaningfully classify optimal child nodes correctly;\\      Achieve better optimality gap.\end{tabular} &
		\begin{tabular}[c]{p{0.17\textwidth}}Find better solutions on   real-world sized instances;\\      A learned policy does not require extra training data.\end{tabular} \\ \midrule
		Limitation(s) &
		Treat every node of a B\&B   tree equivalently, ignoring the uniqueness of each node's feature during   B\&B run. &
		\begin{tabular}[c]{p{0.17\textwidth}}No children selection   policy;\\      Problem dependent, training data should be of the same type;\\      Cannot obtain a certified optimality within the time or node limitation.\end{tabular} &
		\begin{tabular}[c]{p{0.17\textwidth}}Only consider the optimality gap   as the rank criterion,ignore the depth;\\      k-best solutions do not have a optimality gap bound;\\      Assume the optimum was known.\end{tabular} &
		\begin{tabular}[c]{p{0.17\textwidth}}Rely on the provided   solutions;\\      No adjusting in terms of runtime and solution quality.\end{tabular} \\ \bottomrule
	\end{tabular}
\end{table}

\begin{figure}
	\centering
	\includegraphics[width=0.9\textwidth]{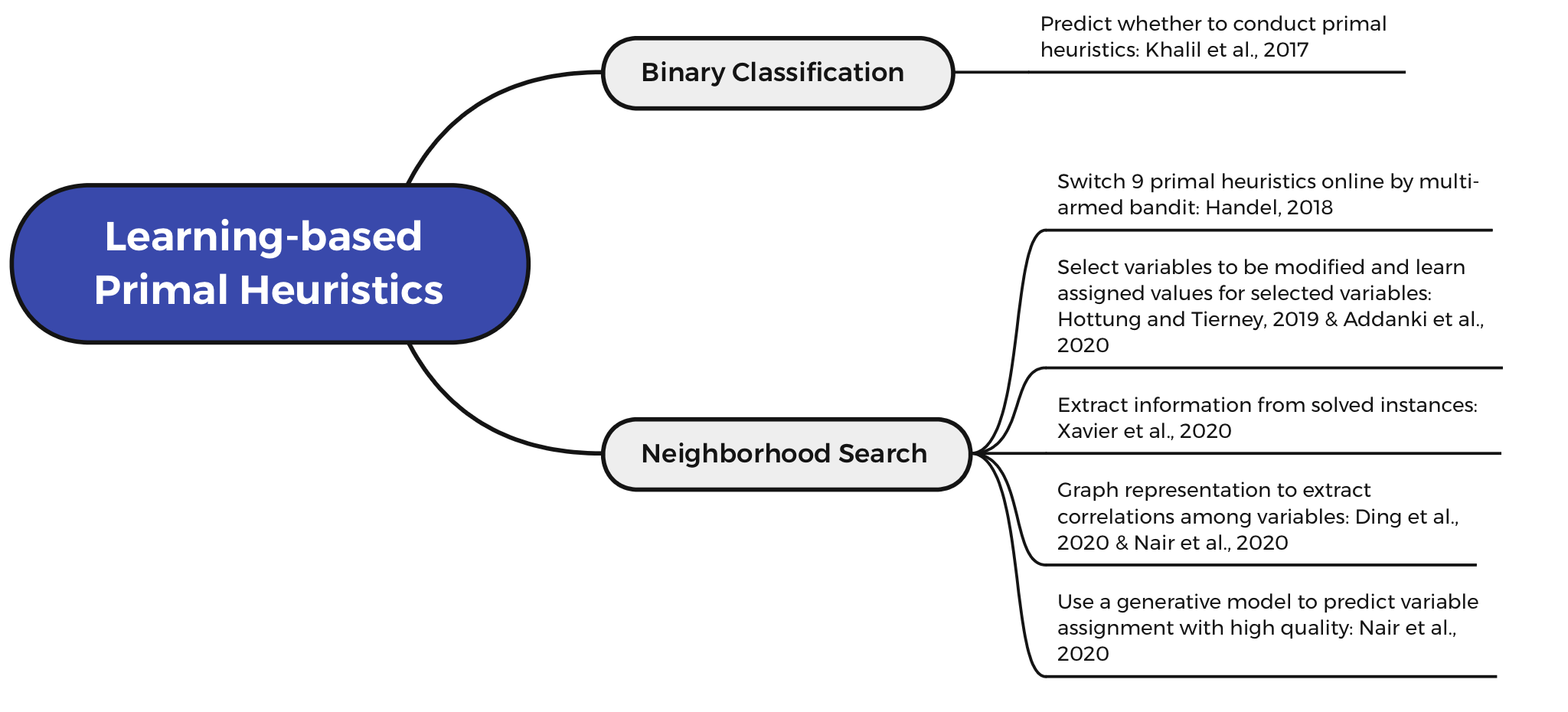}
	\caption{Main studies on learning-based primal heuristics and their main contributions}
	\label{LPH}
\end{figure}

A relatively large amount of literature focuses on embedding learning methods into primal heuristics, including but not limited to \cite{khalil2017learning,hendel2018adaptive,hottung2019neural,addanki2020neural,song2020general,xavier2020learning,ding2020accelerating,nair2020solving}. Khalil et al. \cite{khalil2017learning} used binary classification to predict whether a primal heuristic would succeed at a given node. It is the first systematic attempt at optimizing the use of heuristics in tree search. This work boosts the primal performance of SCIP, even on instances for which experts already finetune the solver. However, the learning rules can be refined, taking into account the running time left for the solver. Hendel \cite{hendel2018adaptive} formulated a multi-armed bandit approach to learn to switch nine primal heuristic strategies online. Hottung and Tierney \cite{hottung2019neural}, Addanki et al. \cite{addanki2020neural} and Song et al. \cite{song2020general} adopted learning methods to improve the neighborhood search to improve the primal performance. The learning methods were adopted to either select the variables to be modified or assign new values to an already selected subset of variables. However, those methods require at least a feasible point as input. Xavier et al. \cite{xavier2020learning} proposed three different learning models to effectively extract information from previously solved instances in order to significantly improve the computational performance of MILP solvers when solving similar instances. The first model 
was designed to predict which constraints should be initially added to the relaxation and which constraints can be safely omitted. The second model proposed using k-nearest neighbors to set the values only for the variables with high confidence and to let the MILP solver determine the values for the remaining variables as a good warm start point. The third model develops an ML model for finding constraints to restrict the solution without affecting the optimal solution set. These three models boost the speed to find a solution with optimality guarantees. The main limitation of this work is that a large number of solved instances must be available for the third model. In addition, it requires the problem to be solved in the future sufficiently similar to the past samples. Unlike the above learning primal heuristic works \cite{khalil2017learning,hendel2018adaptive,hottung2019neural,addanki2020neural,song2020general,xavier2020learning}, Ding et al. \cite{ding2020accelerating} and Nair et al. \cite{nair2020solving} did not limit to problems of certain types and built a tripartite graph representation to extract correlations among variables, constraints and objective function without human intervention. Nair et al. \cite{nair2020solving} posed the problem of predicting variable assignments as a generative modeling problem, which provides a principled way to learn on all available feasible assignments and also generates partial assignments at test time. However, this training method requires GPU.
Fig. \ref{LPH} demonstrates the main contributions of the above literature.

\subsection{Learn to Prune}
When using the B\&B algorithm to solve large-scale systems, the number of non-pruning nodes by traditional pruning methods is still quite massive. As a result, it either costs huge computational consumption to solve the problem, or a possible good solution cannot be obtained within the time limit. Therefore, the question arises of whether we could use some learning methods to effectively prune a large set of unexplored nodes as early as possible and achieve a solution with an optimality gap guarantee. There are mainly three ways to accelerate the pruning policy by learning. One is learning whether to discard or expand the selected node given the current progress of the solver once the popped node is not fathomed, such as \cite{he2014learning}. The second is using learning methods to achieve better bounds to accelerate the pruning process, such as \cite{hottung2020deep}. The last one is trimming the neural network by cutting off redundant connections between weights or neurons of adjacent layers and attaching fine-tuning subsequently to the pruned model for improving the performance, such as Han et al. \cite{han2015deep}. Note that the third method is not a particular method for the B\&B algorithm since it can be applied to simplify the problem model if it implements the neural network. In this survey, we will add more references related to the first two ways as follows. We call the first method as learning-based node pruning and the second one learning-based bounds.

\subsubsection{Learning-based Node Pruning}
Traditional nodes pruning methods only prune nodes if one of the three cases is met as summarized in \ref{s2.4}. Nowadays, learning methods are adopted to prune the non-optimal nodes to accelerate the B\&B algorithm.  He et al. \cite{he2014learning} treated the node pruning policy as a binary classifier that predicts an action in $\{prune, expand\}$. The classifier takes a feature-vector description of the current state and attempts to predict the current action to mimic the expert action, which prunes a node when it does not belong to the optimal branch. This method learns when to leave an unpromising area and when to stop for a good enough solution; thus, it obtains a smaller gap solution using less time than SCIP. However, when the network scale is large, the computational complexity of \cite{he2014learning} is still high since the state space is quite ample. Shen et al. \cite{shen2019lorm} made great efforts to improve the computational efficiency over \cite{he2014learning} by exploiting the structure of the B\&B algorithm and problem data. The authors also proposed self-learning--one kind of transfer learning method without labeling, to address the task mismatch issue by relying on only a few additional unlabeled training samples. As a result, this method requires few training samples and shows effectiveness on acceleration compared to the algorithm in \cite{he2014learning}. Unfortunately, the above two methods are redundant to learn to prune the fathomed nodes. To cope with this difficulty, Lee et al. \cite{lee2019learning} proposed to simplify the learning task by keeping the traditional prune policy and learning an auxiliary prune policy to reinforce it.  Different from only keeping the top solution and pruning the other non-optimal nodes in the above literature, Yilmaz and Smith \cite{yilmaz2021study} kept the best $ k $ solutions. Empirical results on five MIP datasets indicate that this policy leads to solutions significantly more quickly than the state-of-the-art precedent in the literature. In addition, with the same time limitation requirement, this algorithm can achieve a better optimality gap.

\subsubsection{Learning-based Bounds}
Since the MILP is hard to solve, relaxed convex subproblems are solved as the lower bounds. More nodes can be pruned if the lower bound can be determined much closer to the original subproblem.  Hottung et al. \cite{hottung2020deep} used a DNN to heuristically determine the lower bounds. The authors then proposed three pruning functions to determine whether to prune certain branches or nodes based on the output. On the other hand, if the upper bound is loose, which is often much higher than the actual objective, then few branches of the search tree can be pruned. In the traditional B\&B algorithm, one usually updates the upper bound along with diving the tree. However, if a tighter upper bound can be obtained faster, more subproblems can be pruned. The primal heuristics \cite{khalil2017learning,hendel2018adaptive,hottung2019neural,addanki2020neural,song2020general,xavier2020learning,ding2020accelerating,nair2020solving} which aim to find a possible solution as fast as possible give a possible solution with a high quality upper bound; thus they will also have a significant improvement to accelerate the pruning speed. Some special learning-based upper bounds are derived for some particular MILP problems, such as decision trees \cite{aglin2020learning} by Again et al. and informative path planning \cite{binney2012branch} by Binney and Sukhatme.

\subsection{Learn to Cut}
Existing traditional theoretical analysis limit us to understand and address cutting-plane selections, and they have so far failed to help in practical cutting-plane selection~\cite{2018Theoretical} directly. Applying learning to cut generation and selection has the potential to improve the solving MILPs and offer help in understanding and tackling related issues.

\subsubsection{Reinforcement Learning in Cut Selection}
Tang et al.~\cite{pmlr-v119-tang20a} used reinforcement learning to enhance the performance of heuristics. It formulated the process of sequentially selecting cutting planes as a MDP. At iteration $t$, the numerical representation of the state is $s_t=\{ \mathcal{C}^{(t)}, c, x_{LP}^{*}(t), \mathcal{D}^{(t)} \}$, where $\mathcal{C}^{(t)} = \{a_{i}^{T} \leq b_{i}\}_{i=1}^{N_t}$ is the feasible region with $N_{t}$ constraints of the current LP, $c$ is the parameter of objective function. Solving this new LP yields an optimal solution $x_{LP}^{*}(t)$ and the set of candidate Gomory's cuts $ \mathcal{D}^{(t)}$. Thus, the action space at iteration $t$ is $\mathcal{D}^{(t)}$. After taking an action, which is adding one inequality $e_{i}^{T}x \leq d_{i}$ from $\mathcal{D}^{(t)}$, the new feasible region becomes $\mathcal{C}^{(t)} = \mathcal{C}^{(t)} \cup \{e_{i}^{T}x \leq d_{i}\}$. Then $x_{LP}^{*}(t+1)$ and $\mathcal{D}^{(t+1)}$ can be computed, and the new state $s_{t+1}=\{ \mathcal{C}^{(t+1)}, c, x_{LP}^{*}(t+1), \mathcal{D}^{(t+1)} \}$ is determined. The gap between objective values of these LP solutions $r_{t}=c^{T}x^{*}_{LP}(t+1) - c^{T}x^{*}_{LP}(t)$ is the reward for the RL agent at iteration $t$, encouraging the agent to approach the optimal integer solution as fast as possible. The trained policy specifies a distribution over the action space $\mathcal{D}^{(t+1)}$ at a given state $s_t$. To make the policy agnostic to the ordering among the constraints, an attention network was adopted. Besides, to enable the network to handle IP instances with various sizes, LSTM with hidden states was utilized, which encoded all information in the original inequalities. Concerning four classes of IP instances: Packing, Production Planning, Binary Packing, and Max-Cut, experiment results illustrate the performance of the RL agent from five perspectives: efficiency of cuts, integrality gap closed, generalization properties, impact on the efficiency of branch-and-cut, and interpretability of cuts. Compared to some commonly used human-designed heuristics: Random, Max Violation, Max Normalized Violation, and Lexicographical Rule, the RL agent needs the least number of cuts. For some larger-scale instances with sizes close to $5000$, the agent can close the highest fraction of integrality. Although trained on smaller instances, the RL agent can still have extremely competitive performance on $10$ times larger test problems. Moreover, even if the agent is trained on small packing problems, it can be applied to $10$ times larger graph-based maximum-cut problems, indicating its great generation properties. Compared to the pure B\&B method, this learning-based cutting-plane selection reduces the number of expanded nodes significantly. However, the running time of the RL policy is not improved substantially compared with other methods since running the policy is somehow costly. 

\subsubsection{Ranking-based Cut Selection}
With the aim to reduce the total running time, Huang et al.~\cite{huang2021learning} proposed a new metric, named Problem Solvability Improvement, to measure the quality of the selected cut subset. Since obtaining the problem solvability is infeasible in practice, this metric was substituted with the reduction ratio of solution time of selecting a cut subset. A score function that measures this quality was learned as a rank formulation which labels the top-ranked cuts as positive and the remaining cuts as negative because the main goal is to differentiate good cuts from poor cuts, rather than to rank all candidate cuts. Different from Khalil et al.~\cite{Khalil_2016}, labels for individual cuts are not easy to obtain. Hence, Multiple Instance Learning (MIL) was capitalized on to tackle this issue. Instead of requiring instances labeled individually, in Babenko~\cite{babenko2008multiple}, MIL receives a set of labeled bags, each including several instances. In the binary classification, a bag would be labeled negative when all instances in it are negative. Otherwise, a bag would be labeled positive as long as there is at least one positive instance in it. This technique rather fits the scenario of cut selection since the label assignment is determined by more than one instance, and the effect of an individual cut is very imperceptible for large-scale problems. To better generalize the mode, 14 problem-specific atomic features adapted from~\cite{Khalil_2016} were designed, including statistics of cut and objective coefficients, support, normalized violation, distance, parallelism, and expected improvement. Experimental results demonstrate the power of this ranking-based cut selection policy in terms of the quality, generalization ability, and performance on large-scale real-world tasks. The cut ranking policy has reduced the solving time and the number of nodes more significantly than human-designed heuristics, and shown higher stability. Besides, this learned policy has a certain generalization ability on instances with different sizes, structures. However, it may have inferior performance when encountering a large range of coefficients. It should be noted that this work firstly applied the learning-based cut selection strategy for large-scale MILPs with more than $10^{7}$ variables and constraints. For real-world production planning problems, this method improves the efficiency of Huawei's industrial MILP solver and solves problems without loss of accuracy, speeding up the ratio of $14.98\%$ and $12.42\%$ in the offline and online setting, respectively.

Besides, several works have applied learning to cut generation in nonlinear programming, such as  Dey et al.~\cite{dey2021cutting} and Baltean-Lugojan et al.~\cite{baltean2018selecting}. This topic is beyond our scope, and we will not talk about related works in detail.

\section{Future Work Direction and Methodology}\label{s5}
In this section, we review some of the algorithmic concepts previously introduced by taking into account their limitations. Future work directions to cope with their limitations and possible methodologies are included in this section.

\begin{figure}[!t]
	\centering
	\includegraphics[width=0.5\textwidth]{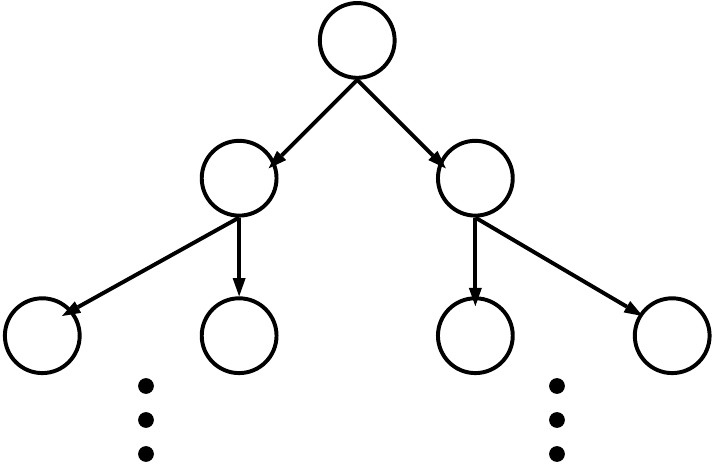}
	\caption{Illustration that depth-first would cost redundant effort to solve nonoptimal nodes. (If the optimal node belongs to the right side tree, it would cost a lot of effort but is meaningless to search the whole left side tree.)}
	\label{fd}
\end{figure}

\subsection{Online Tuning Learning Methods}
In Section \ref{s4.2.1}, we reviewed a work using learning methods to obtain a suitable tuning parameter, which balances the best-first and depth-first in the B\&B run. However, this work ignores the uniqueness of each node during the B\&B run. For example, it should count more on the quality of the achieved solution for the root node since achieving a solution with high quality would significantly reduce the searching space. Otherwise, it should make a redundant effort to solve nonoptimal nodes, as illustrated in Fig. \ref{fd}. Therefore, it would be preferable for the root node to assign more weight to the best-first policy.  On the other hand, satisfying the feasibility constraints would be more critical for the near leaf nodes since the quality can be adjusted with just a few backward exploitations. From the above analysis, it would be interesting to consider the nodes' features (such as depth) and propose an online tuning learning method to adjust the tradeoff along with the B\&B run.

Moreover, we can improve the estimate in \cite{sabharwal2012guiding} using the evaluation criteria summarized in Section \ref{s2.3.3}. Instead of using the normalized relaxed LP objective value as the estimate of this node in \cite{sabharwal2012guiding}, we can take into account the feasibility quality along with the optimality quality. We can further introduce a parameter to tune the importance of the feasibility quality versus the optimality quality and learn this parameter online.

\subsection{Extension to Nonbinary Classifications and Adapt Learning Strategies at a Given Time}
Most literature in this survey does not consider the time required to adjust the optimal learning strategy. For example, in Section \ref{s4.2.2}, existing literature \cite{khalil2017learning} adopts binary classification to learn whether primal heuristics will succeed at a given node. However, whether to conduct primal heuristics might also be influenced by the quality constraints and time limitation for implementation, and all those can not be included in the binary classification.

\begin{figure}[!t]
	\centering
	\includegraphics[width=0.5\textwidth]{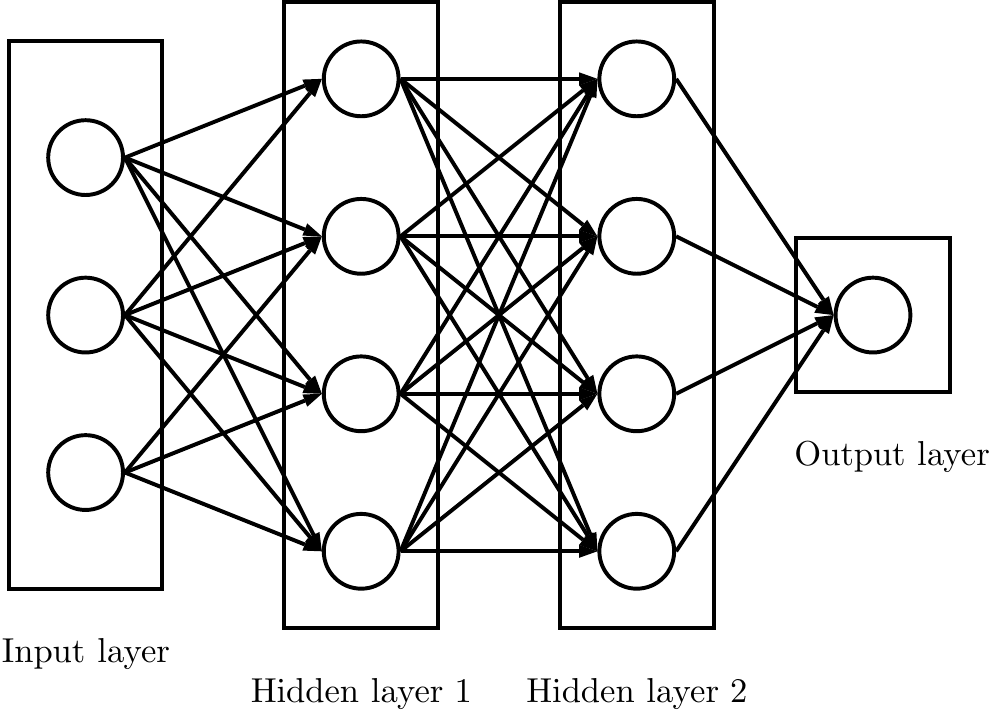}
	\caption{Demonstration of 3-layer multi-layer perceptron}
	\label{MLP}
\end{figure}

Inspired by works \cite{nair2020solving,shen2019lorm} and so on, we could expand the output to indicate the probability of whether to conduct a primal heuristics (or other problems which has binary classes) with problem features $( A,b,c)$ or the extracted feature by GCNN \cite{gasse2019exact,gupta2020hybrid,ding2020accelerating,nair2020solving} as shown in Fig. \ref{bigraph} and \ref{gnn}. The learning task is to approximate the probability distribution. We employ an L-layer multi-layer perceptron, a type of neural network to learn the mapping from the input feature vector to the output indicating the probability of each class, as shown in Fig. \ref{MLP}.  We then adopt the supervised learning, or Data Aggregation \cite{ross2011reduction} to learn the model parameters by minimizing the weighted loss function. The time limitation could be considered to tune the threshold. For example, in the standard classification, if the probability is larger than $ 0.5 $ of belonging to the first class, we cluster this input to the first class; otherwise, it belongs to the second class. Suppose the first-class means conducting primal heuristics, decreasing the threshold leads to a larger exploration space and obtains a feasible solution with higher quality. We could iteratively decrease the threshold to satisfy the quality constraints and time limitations and obtain the learned parameter adaptively.

\subsection{Reduce Training Samples}
A common limitation of learning-based algorithms in this survey is that in order to have good predictors, a large number of solved instances must be available, and the solved instances in most literature require to be of the same type as the problem we expect to solve in the future. However, when the problem scale becomes quite large, generating the training data is costly, let alone training the network by the data.

\begin{figure}[!t]
	\centering
	\includegraphics[width=0.6\textwidth]{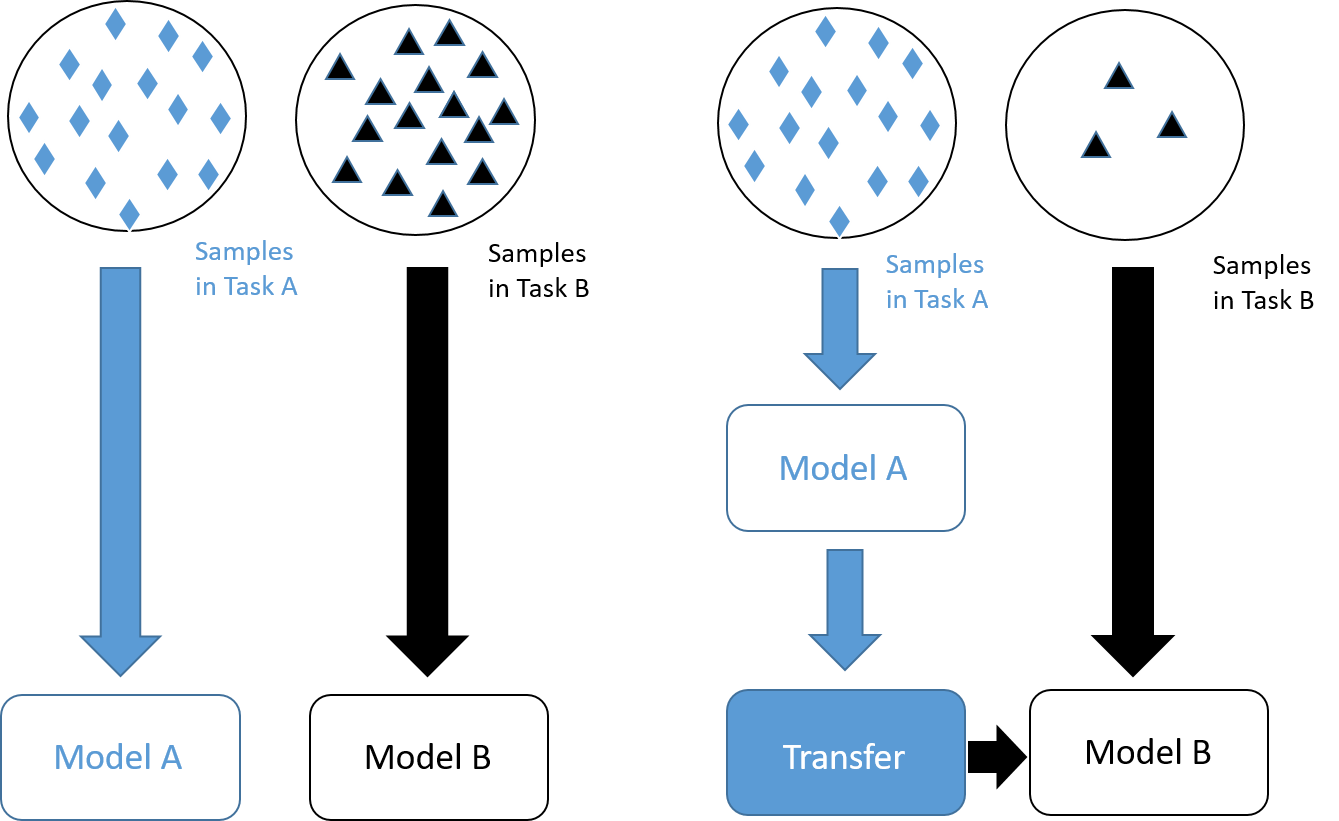}
	\caption{Demonstration of the difference between transfer learning and traditional ML.}
	\label{TL}
\end{figure}

In order to address this limitation, there are two possible directions. One is generating the training data along with the problem-solving progress. Dataset Aggregation (DAGGER), which was proposed by Ross et al. \cite{ross2011reduction}, is an iterative algorithm that trains a deterministic policy that achieves good performance guarantees under its induced distribution of states. Pseudocode for DAGGER is shown in Algorithm \ref{al3}, where $\pi^{*}$ is the presence of the expert in DAGGER and $\beta_{i} \in [0,1]$ represents the trust of the expert decisions. The other direction is using transfer learning to exploit the pre-trained neural network and train a new model with only a few additional samples. Fig. \ref{TL} demonstrates the transfer learning method. The essential advantage is that with transfer learning, the new task can be trained with fewer additional training samples than the traditional one since some information can be achieved by transferring the knowledge from the old task into the new task.
\begin{algorithm}[htb]
	\caption{DAGGER Algorithm} 
	\label{al3}
	\begin{algorithmic}[1]
		\State Initialization: $ D\leftarrow\emptyset $,  and pruning policy $ \hat{\pi}^{(0)}$.
		\For {$ i=0:N $}
		\State Let $ \pi^{(i)}\leftarrow\beta_{i}\pi^{*}+(1-\beta_{i})\hat{\pi}^{(i)} $
		\State Sample $T-$step trajectories using $ \pi^{(i) } $.
		\State Collect dataset  $  D_{i}=\{(s,\pi^{*}(s))\} $ of visited states by $ \pi^{(i)} $.
		\State Aggregate datasets  $ D\leftarrow D\cup D_{i} $.
		\State Train the policy $ \pi^{(i+1)} $ based on $ D $.
		\EndFor
		\State \Return best $\hat{\pi}^{(i)}  $ on validation.
	\end{algorithmic}
\end{algorithm}

\subsection{Extension to Specific Joint Bilinear Problems and  Discrete Network Design Problems}
A wide range of resource management problems in networks can be formulated as jointly bilinear problems. For example, a concave piecewise linear network flow problem (CPLNFP) is equivalent to a jointly constrained bilinear program \cite{nahapetyan2007bilinear}. Let $ G(N,V) $ represent a network where $ N $ and $ V $ are the sets of nodes and arcs, respectively. The general form of the CPLNFP is 
\begin{equation}\label{e14}
\begin{split}
&\min\limits_{x}\sum\limits_{a\in A} f_{a}(x_{a}),\\ 
&s.t. \, x\in X, x_{a}\in[\lambda_{a}^{0},\lambda_{a}^{n_{a}}], \forall a\in V,
\end{split}
\end{equation}
where $ X $ is a convex set representing the constraints on $ x $, and $ f_{a}(x_{a}) $ are piecewise linear concave functions, i.e.,
\begin{equation}
f_{a}(x_{a}) =\left\lbrace\begin{array}{l}
c_{a}^{1}x_{a}+s_{a}^{1}(=f_{a}^{1}(x_{a})), x_{a}\in[\lambda_{a}^{0},\lambda_{a}^{1}),\\ 
c_{a}^{2}x_{a}+s_{a}^{2}(=f_{a}^{2}(x_{a})), x_{a}\in[\lambda_{a}^{1},\lambda_{a}^{2}),\\ 
\colon\\ 
c_{a}^{n_{a}}x_{a}+s_{a}^{n_{a}}(=f_{a}^{n_{a}}(x_{a})), x_{a}\in[\lambda_{a}^{n_{a}-1},\lambda_{a}^{n_{a}}],
\end{array}  \right. 
\end{equation}
with $ c_{a}^{1}>c_{a}^{2}>\cdots>c_{a}^{n_{a}} $. The solution of the above problem \eqref{e14} can be easily constructed from a global solution of the following jointly bilinear problem \cite{nahapetyan2007bilinear}
\begin{equation}
\begin{split}
&\min\limits_{x,y}\sum\limits_{a\in A}\sum\limits_{i=1}^{n_{a}} f_{a}^{i}(x_{a})y_{a}^{i},\\ 
&s.t. \, x\in X,\, \sum\limits_{i=1}^{n_{a}}\lambda_{a}^{i-1}y_{a}^{i}\leq x_{a}\leq\sum\limits_{i=1}^{n_{a}}\lambda_{a}^{i}y_{a}^{i},\, \sum\limits_{i=1}^{n_{a}} y_{a}^{i}=1,\, x_{a}\geq 0, y_{a}^{i}\geq0.
\end{split}
\end{equation}

A general jointly bilinear constrained program is
\begin{equation}
\begin{split}
&\min\limits_{x,y}f(x)+x^{\top}Ay+g(y),\\ 
&s.t. \, (x,y)\in S\cap\Omega,
\end{split}
\end{equation}
where $ x\in\mathbb{R}^{p} $ and $  y\in\mathbb{R}^{q} $ are decision variables; $ A $ is a given $ p\times q $ matrix; $ f $ and $ g $ are given functions which are convex over $ S\cap\Omega $; $ S $ is a closed, convex set; and $ \Omega=\{(x,y): l\leq x\leq L, m\leq y\leq M\} $. It is shown in \cite{al1983jointly} that the optimal solution of a general jointly constrained bilinear program belongs to the boundary of the feasible region.

Different from MILP iteratively constructing a binary search tree, the B\&B algorithm for jointly bilinear constrained program generates a four-branch search tree iteratively by splitting a selected rectangle into four subrectangles. In addition, since the variables are changed from discrete integers to continuous intervals, the stopping criterion is also changed. From the above analysis, the complexity of this problem increases dramatically. One possible future work is to use learning methods to accelerate the branch, select, pruning, and cutting procedure in joint bilinear problems, which can be used to accelerate solving the large-scale discrete network design problems.

\section{Conclusions} \label{s6}
In this paper, we have surveyed the existing literature studying different approaches and algorithms for the four critical components in the general B\&B algorithm, namely, branching variable selection, node selection, node pruning, and cutting-plane selection. However, the complexity of the B\&B algorithm always grows exponentially with respect to the increase of the decision variable dimensions. 

As the problem dimension size increases in real applications, the traditional B\&B algorithm is not able to obtain a certified solution within the time limit. In order to improve the speed of B\&B algorithms, learning techniques have been introduced in this algorithm recently (mainly in the past decade). We further surveyed how machine learning can be used to improve the four critical components in B\&B algorithms. In general, a supervised learning method helps to generate a policy that mimics an expert but significantly improves the speed. An unsupervised learning method helps choose different methods based on the features. In addition, models trained with reinforcement learning can beat the expert policy, given enough training and a supervised initialization. Detailed comparisons between different algorithms have been summarized in our survey.

Although most of the approaches we discussed in this paper speed the run time and achieved better gaps with respect to the optimal solutions. Most learning approach does not consider the features generated from the B\&B run; that is to say, they view all the nodes equally and then derive a deterministic approach based on the problem-dependent features. To further improve the speed of the B\&B algorithm, one possible direction is finding an online tuning method based on the features generated from the B\&B run. Meanwhile, another major limitation is that most literature in this survey does not take into account the time requirement and the quality bound. The learning algorithm, which is adaptive with respect to the time and quality requirement, is worth studying. Apart from that, for large-scale problems, generating training samples would also cost significant computational effort. Reducing the training samples is another big challenge. Last, the discrete network design problems do not fall into the MILP problems but can be viewed as a joint bilinear problem. How to extend the above learning methods to this new type of problem is also a possible topic. Finally, we discuss some possible methodologies on these future directions.

\bibliographystyle{IEEEtran}
\bibliography{bbreference}
\end{document}